%% file: uist24-8.tex
\documentclass[sigconf]{acmart}

\input{_defines}

\AtBeginDocument{%
  }

\copyrightyear{2024}
\acmYear{2024}
\setcopyright{rightsretained}
\acmConference[UIST '24]{The 37th Annual ACM Symposium on User Interface
Software and Technology}{October 13--16, 2024}{Pittsburgh, PA, USA}
\acmBooktitle{The 37th Annual ACM Symposium on User Interface Software and
Technology (UIST '24), October 13--16, 2024, Pittsburgh, PA, USA}
\acmDOI{10.1145/3654777.3676330}
\acmISBN{979-8-4007-0628-8/24/10}

\acmSubmissionID{2943}

\begin{document}

\title[TouchInsight: Uncertainty-aware Rapid Touch and Text Input for Mixed Reality from Egocentric Vision]{TouchInsight: Uncertainty-aware Rapid Touch and Text Input\\for Mixed Reality from Egocentric Vision}

\author{Paul Streli}
\orcid{0000-0002-3334-7727}
\affiliation{%
\institution{Meta Reality Labs \\
    \city{Redmond}\state{WA}\country{USA}~\&}
 \institution{Department of Computer Science \\ ETH Zürich
    \country{Switzerland}}
}

\author{Mark Richardson}
\orcid{0009-0005-1603-8374}
\affiliation{
 \institution{Meta Reality Labs}
 \city{Seattle}
 \state{WA}
 \country{USA}
}

\author{Fadi Botros}
\orcid{0009-0003-6389-7126}
\affiliation{
 \institution{Meta Reality Labs}
 \city{Redmond}
 \state{WA}
 \country{USA}
}

\author{Shugao Ma}
\orcid{0000-0002-4986-2221}
\affiliation{
 \institution{Meta Reality Labs}
 \city{Redmond}
 \state{WA}
 \country{USA}
}

\author{Robert Wang}
\orcid{0000-0002-9298-7337}
\affiliation{
 \institution{Meta Reality Labs}
 \city{Redmond}
 \state{WA}
 \country{USA}
}

\author{Christian Holz}
\orcid{0000-0001-9655-9519}
\affiliation{%
 \institution{Department of Computer Science \\ ETH Zürich
    \country{Switzerland}}
}

\renewcommand{\shortauthors}{Streli et al.}

\input{sections/0-abstract}


\input{sections/_teaser.tex}


 
\maketitle

\input{sections/1-introduction}

\input{sections/2-related_work}

\input{sections/3-method}

\input{sections/4-implementation}

\input{sections/5a-location}

\input{sections/5b-textentry}

\input{sections/7-limitations}

\input{sections/8-conclusion}

\begin{acks}
We thank Yangyang Shi, Bradford Snow, Pinhao Guo, and Jingming Dong for helpful discussions and comments, as well as the participants of our user studies.
\end{acks}

\balance
\bibliographystyle{ACM-Reference-Format}
\bibliography{references.bib}

\end{document}

%% file: _defines.tex
\usepackage{bm}
\usepackage{mathtools}
\usepackage{multirow}
\usepackage{color}
\usepackage{enumitem}
\usepackage[utf8]{inputenc}
\usepackage{setspace} 
\usepackage{siunitx} 
\usepackage{geometry}

\def\projectname{Touch\-Insight}
\def\optitrack{\textsc{OptiTrack}}
\def\umetrack{\emph{UmeTrack}}
\def\midair{\textsc{Midair}}
\def\greedy{\textsc{GreedySurf}}
\def\beam{\textsc{BeamSurf}}

\def\twitteriv{\textsc{TwitterIV}}
\def\twitteroov{\textsc{TwitterOOV}}

\def\blockone{\textsc{Block~1}}
\def\blocktwo{\textsc{Block~2}}
\def\blockthree{\textsc{Block~3}}
\def\blockoov{\textsc{Block~OOV}}

\newcommand{\anova}[6]{{\small [$F_{#1,#2}$$=$\allowbreak$#3$, $p$$#4$\allowbreak$#5$,$\eta_{p}^{2}$$=$\allowbreak$#6$]}}
\newcommand{\artanova}[5]{{\small [$F_{#1,#2}$$=$\allowbreak$#3$, $p$$#4$\allowbreak$#5$]}}
\newcommand{\pval}[2]{{\small ($p#1#2$)}} 
\newcommand{\pvall}[2]{{\small $p#1#2$}} 

\DeclareMathOperator*{\argmax}{argmax}


%% file: sections/0-abstract.tex
\begin{abstract}
While passive surfaces offer numerous benefits for interaction in mixed reality, reliably detecting touch input solely from head-mounted cameras has been a long-standing challenge.
Camera specifics, hand self-occlusion, and rapid movements of both head and fingers introduce considerable uncertainty about the exact location of touch events.
Existing methods have thus not been capable of achieving the performance needed for robust interaction.

In this paper, we present a real-time pipeline that detects touch input from all ten fingers on any physical surface, purely based on egocentric hand tracking.
Our method \emph{\projectname} comprises a neural network to predict the moment of a touch event, the finger making contact, and the touch location.
\projectname{} represents locations through a bivariate Gaussian distribution to account for uncertainties due to sensing inaccuracies, which we resolve through contextual priors to accurately infer intended user input.

We first evaluated our method offline and found that it locates input events with a mean error of 6.3\,mm, and accurately detects touch events ($F_1=0.99$) and identifies the finger used ($F_1=0.96$).
In an online evaluation, we then demonstrate the effectiveness of our approach for a core application of dexterous touch input: two-handed text entry.
In our study, participants typed 37.0 words per minute with an uncorrected error rate of 2.9\% on average.

\end{abstract}

%% file: sections/_teaser.tex
\begin{teaserfigure}
    \centering
    \vspace{-3mm}
    \includegraphics[width=\linewidth]{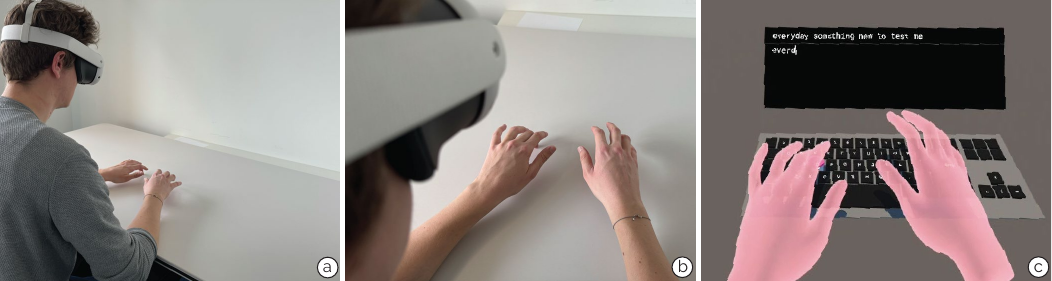}
    \vspace{-6mm}
    \caption{We present a novel method for detecting touch input on surfaces from egocentric hand tracking alone.
(a)~Here, a user is interacting in mixed reality through touch on a desk, which provides haptic feedback.
(b)~Our method senses input events from egocentric views---a challenging task, as hand self-occlusion causes sensing uncertainty about exact touch locations.
Our learning-based method explicitly models these uncertainties and resolves them as part of a probabilistic framework, accounting for user behavior and context to enable rapid and dexterous text input on a virtual surface keyboard (c).
    }
    \Description{(a) The figure shows a user engaged with a virtual reality interface. The individual is equipped with an MR headset and is seated at a desk, interacting with a surface that delivers passive haptic feedback. (b) The figure depicts a close-up view of hands positioned above a flat surface, highlighting the intricacies of sensing touch from a first-person perspective. The hand placement suggests typing on an invisible (virtual) interface, with the potential difficulty of precise touch location detection due to the fingers obscuring each other. (c) The figure shows two virtual tracked hands floating above a virtual keyboard in VR. The user transcribes a phrase on the keyboard with a QWERTY layout and black keys.
    }
    \label{fig:teaser}
\end{teaserfigure}

%% file: sections/1-introduction.tex
\section{Introduction}

Direct interaction has become a preferred form of input on mixed reality (MR) headsets~\cite{directtouch}.
Users can intuitively manipulate virtual objects and interfaces using their hands and fingers, enabled by advancements in real-time egocentric hand tracking~\cite{han2020megatrack,han2022umetrack}.
While MR interaction has so far been designed for mid-air interaction, recent research has demonstrated the benefits of moving these interactions to surrounding \emph{passive} surfaces, such as for improved input control and performance~\cite{passivefeedbackeffectpresencetaskperformance2010abad, henderson2010exploring}.
Surfaces provide haptic feedback and an opportunity for rest, and thus make the interaction more comfortable and avoid fatigue during prolonged use~\cite{ismar2022-comfortableUIs}.

Users are well-acquainted with touch interaction on physical objects or surfaces in the real world.
Arguably, the success of smartphones and tablets is due to their focus on this input modality, even for intricate tasks such as text entry~\cite{vertanen2019velociwatch}.
Transferring this mode of interaction to MR systems is difficult, however; following their transition to built-in tracking without the need for stationary cameras, touch input needs to be \emph{inferred} from camera observations.

Accurately inferring touch events from egocentric views is considerably challenging.
Challenges include hand self-occlusion that results in uncertainty about the exact locations of fingers, detecting precise moments of physical contact, and obtaining the corresponding touch input locations.
Previous research on vision-based contact recognition has thus far required fingers and touch surfaces to be clearly visible~\cite{dupre2024tripad, grady2022pressurevision, grady2024pressurevision++}.
To enable more dexterous and rapid input, as is needed for typing, additional instrumentation of the hands~\cite{chi2022-taptype, zhang2022typeanywhere, telemetring} or external motion capture has been required~\cite{richardson2020decoding} to achieve practical accuracy for downstream tasks.

In this paper, we present \textit{\projectname}, a method for recognizing touch input on any physical surface for all ten fingers from the egocentric views of moving head-mounted cameras.
Our method identifies moments of touch events and estimates their input locations.
The key novelty is our explicit model of the uncertainty inherent to the sensing pipeline, in addition to user behavior.
This allows our method to robustly infer user intentions even for quick touch input and when hands are self-occluded in egocentric vision. 
Our method has particular implications for two-handed surface typing, which we use as a challenging task in our online evaluation.


\subsection*{Inferring touch from uncertain input observation}

\begin{figure}[t]
    \centering
    \includegraphics[width=\columnwidth]{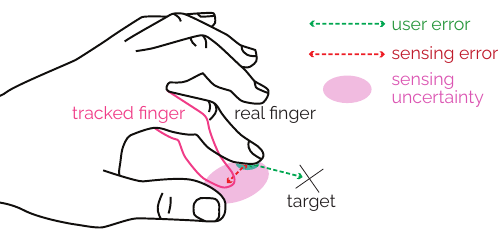}
    \caption{For a single touch input event, the offset between the target and the finger endpoint can be considered as \textit{user error}.
    Due to inaccuracies in hand tracking, the endpoint of a tracked finger might additionally deviate from the actual finger location.
    We refer to this as \textit{sensing error}, which introduces \textit{sensing uncertainty} about the touch location.}
    \label{fig:errorvisualisation}
    \Description{The image visualizes the concept of different touch input errors for MR hand tracking. It consists of a line drawing of a hand from an overhead perspective, illustrating the difference between where a finger is tracked versus its actual position when targeting a specific point. There are three main elements: Tracked Finger: Indicated by a pink line, it shows the pose of the index finger as estimated by the hand tracking system. Real Finger: Shown by a black line, this represents the actual position of the finger as it moves towards a target. Target: Marked by an 'X', this is the intended point the user tries to touch. Two types of dashed lines indicate errors: User Error: A green dashed line showing the distance between the target 'X' and the actual end position of the real finger, representing the inaccuracy from the user's side. Sensing Error: A red dashed line illustrating the distance between the tracked position of the finger and its real end position, showing the error caused by the hand tracking technology. There is also a shaded oval labeled ``sensing uncertainty'', which highlights the area of ambiguity due to sensing inaccuracies.
    }
\end{figure}

\autoref{fig:errorvisualisation} illustrates the problem we address in this paper: uncertainty estimation across a series of error sources for egocentric vision-based touch detection.
As a user aims to touch a target with their finger (e.g., to activate a command), resulting endpoint finger locations form a Gaussian distribution~\cite{bi2013ffitts, rot_dual_gauss}.
This distribution captures the inaccuracy of touch input due to user behavior (\textit{user error}), reflecting the speed-accuracy trade-off in the human motor system and the absolute precision of finger touch in general~\cite{holz2010generalized,holz2011}.
The second source of uncertainty stems from egocentric hand tracking;
due to tracking inaccuracy in any hand-pose estimator~\cite{han2022umetrack} and reduced visibility due to self-occlusion (see \autoref{fig:teaser}), recovered hand poses can significantly differ from the actual hand configuration (\textit{sensing error}).
This introduces \textit{sensing uncertainty} about the tracked finger endpoints.
Consequently, using these hand poses as input to infer touch events introduces irreducible \textit{aleatoric} uncertainty in the estimated touch locations.

While `user error' as the first source of uncertainty has received much attention in previous work on touch sensing~\cite{bi2013ffitts,holz2010generalized,rot_dual_gauss}, the significance of `sensing error' as the second source has so far been negligible, as touch sensors have high resolution and precision~\cite{streli2021capcontact}.
However, for camera-based touch estimation, the second error source substantially exceeds the first in magnitude, particularly as the camera sensor is moving and both its effective resolution and sampling rate are lower when capturing the user's fingers.

The key novelty of our approach is to explicitly model these two sources of uncertainty through bivariate Gaussian distributions. 
Our method effectively integrates sensing uncertainties with user uncertainties in a closed-form expression that allows us to reason about potential user intentions in a probabilistic command prediction framework based on touch locations.
Implementing our framework, we present a purely vision-based ten-finger text entry system on a surface-aligned virtual keyboard that runs on a standalone mobile MR headset.
Our text entry system fuses the probabilistic touch and uncertainty estimates with per-key touch distributions to obtain a likelihood distribution for the \emph{intended} character over the vocabulary of the keyboard.
We refine text input predictions through the language priors from a 6-gram character language model and resolve decoding errors through beam search to consider additional priors from a trigram word language model.

We evaluated our method in two folds.
In an offline evaluation, we quantified \projectname{}'s accuracy in detecting touch events, achieving an F$_{1}$-score of 0.99 for event detection and 0.96 for correctly identifying the contacting finger, with a mean temporal latency of less than 70\,ms.
We also assessed our network’s accuracy in predicting location, achieving a mean position error of 6.3\,mm while providing meaningful uncertainty estimates.
To determine the practical value and efficacy of our method, we conducted an online evaluation for a text entry task where 12 participants transcribed sentences from the Twitter dataset~\cite{twitteroov}.
Participants achieved a mean text entry rate of 37.0 words per minute (WPM) with an uncorrected error rate (UER) of 2.9\% after five phrases of training.
Our method thereby significantly outperformed index finger-based text entry on a mid-air keyboard (19.7\,WPM with 8.0\% UER) in terms of input performance, task load, and user preference.

\subsection*{Contributions}

In summary, our work makes the following contributions:

\begin{itemize}[leftmargin=*]
    \item a probabilistic command prediction framework to accurately infer intended user input.
    Our framework incorporates uncertainties from both user behavior and sensing inaccuracies.
    
    \item a novel method to detect touch input on physical surfaces based on egocentric hand tracking.
    Given the environment, our neural network identifies contact events ($F_1$=0.99), the specific finger involved ($F_1$=0.96), the timing of the touch event (<3\,ms mean error with an intended offset of 2\,frames at 30\,Hz), and its location (6.3\,mm offset).
    Importantly, \projectname{} simultaneously estimates the inherent uncertainty of the touch location through a bivariate Gaussian distribution.
    
    \item a novel text entry system for ten-finger text input on physical surfaces in mixed reality with passive haptic feedback.
    Our system runs in real-time and solely takes hand images captured by the Quest~3 cameras from an egocentric perspective as input. In an online user study with 12 participants who transcribed sentences from the Twitter dataset~\cite{twitteroov} using our text entry system, participants achieved 37.0\,WPM with an uncorrected error rate of 2.9\%, compared to 19.7\,WPM with 8.0\% UER in mid-air.
\end{itemize}

%% file: sections/2-related_work.tex
\section{Related Work}

Our work is related to touch sensing, text entry in mixed reality and virtual reality (VR), as well as statistical keyboard decoding.

\subsection{Touch interaction in MR/VR}
Vision-based hand pose estimation has made significant progress over the last decade, driven by the advancements in learning-based methods~\cite{mueller2017real,  pavlakos2024reconstructing} and the collection of relevant datasets~\cite{zimmermann2019freihand, garcia2018first, kwon2021h2o}.
Performing hand tracking from an egocentric perspective on MR headsets~\cite{han2020megatrack, han2022umetrack} presents unique challenges.
The perspective causes a high degree of self-occlusion for numerous hand poses~\cite{rgb2hands, mueller2019real}.
State-of-the-art solutions primarily operate on multi-view image sequences captured by wide field-of-view monochrome cameras~\cite{han2020megatrack, han2022umetrack} to obtain scale information and offer a wide tracking volume.
They leverage temporal information to reduce uncertainty in the estimated poses and ensure temporal consistency~\cite{han2020megatrack, han2022umetrack}.

Their performance and maturity have ultimately led to the availability of hand pose tracking in today's commercial mixed reality headsets (e.g., Quest 3~\cite{quest3}, Apple Vision Pro~\cite{visionpro}, 
Magic Leap 2~\cite{magicleap2}).
This enables direct interaction with virtual objects and interfaces~\cite{passivefeedbackeffectpresencetaskperformance2010abad,bowman20083d,luong2023controllers}, which, so far, has been predominantly confined to mid-air~\cite{luong2023controllers}.
However, research has demonstrated the benefits of leveraging real-world physical objects for inputs as proxies~\cite{opportunisticcontrols2010henderson, simeone2015substitutional, cheng2017sparse, he2023ubi}, offering advantages in terms of passive haptic feedback~\cite{cheng2017sparse} and increased input control~\cite{opportunisticcontrols2010henderson}.
Aligning interfaces with physical surfaces improves accuracy, task performance, and agency while decreasing physical exertion~\cite{ismar2022-comfortableUIs}.

Sensing touch on surfaces is challenging.
External tracking systems have been used for interactive research purposes~\cite{cheng2017sparse, wilson2010combining, richardson2020decoding, agarwal2007high}.
For mobile scenarios, prior work has integrated wearable sensors (e.g., acoustic sensors~\cite{Acoustic}, inertial measurement units (IMUs) attached to fingers~\cite{telemetring, Yizheng_IMU, finger_mounted} or wrists~\cite{meier2021tapid, Gupta_acustico, chi2022-taptype}). 
In addition, vision-based solutions have been proposed that employ additional markers~\cite{lee2003arkb} or require active illumination to cast shadows~\cite{liang2023shadowtouch, wilson2005playanywhere}, sense vibrations~\cite{streli2023structured} or depth~\cite{depth_estimate, Dante_vision,harrison2011omnitouch,Imaginary_phone,FarOut,Ian_depth,Xu_latency, mrtouch2018xiao, xiao2016direct}.

To estimate contact from monocular images, alterations of the fingernail~\cite {chen2020estimating, mascaro2004measurement} or object deformations~\cite{chen2020estimating} during press events provide visual cues.
Alternatively, prior work estimates contact by analyzing the intersections between estimated hand meshes and objects~\cite{grady2021contactopt, hasson2019learning}, constrained by the requirement for millimeter-level accuracy.
Changes in object trajectory and required interaction forces~\cite{ehsani2020use, li2019estimating, pham2017hand} offer insights into hand-object interaction but fail with static environment objects like tables and walls.

\citeauthor{grady2022pressurevision}~\cite{grady2022pressurevision, grady2024pressurevision++} proposed a neural network to directly estimate contact regions on surfaces from single RGB images, but their method relies on an external static camera and good visibility of the corresponding fingertips.
\citeauthor{dupre2024tripad}~\cite{dupre2024tripad}'s technique using Oculus Quest Pro hand tracking for index finger input implements a deterministic state-machine to handle tracking inaccuracies but is unsuitable for rapid input events (e.g., typing).

In contrast to prior work, our method, \projectname{}, detects rapid input on passive surfaces from all ten fingers, solely based on egocentric hand tracking, and explicitly models uncertainties about touch locations due to tracking inaccuracies.

\subsection{Probabilistic command and text prediction}
Unlike physical keyboards, soft keyboards add uncertainty regarding the user's intended key selection due to a lack of confirming haptic feedback, occlusion, and `fat finger` error~\cite{holz2010generalized, holz2011}.
This affects user interfaces, requiring keys to be large enough to accommodate noisy user input when processing input deterministically.

Probabilistic input frameworks can offer an alternative solution by accounting for these user uncertainties and continuously inferring a distribution over potential user intentions~\cite{touchml, williamson2006continuous, weir2012user, gaussianprocess}, which are taken into account throughout the interaction sequence~\cite{schwarz2010framework, schwarz2011monte}.
Bayes' theorem has become a cornerstone for handling and resolving these uncertainties~\cite{bi2013bayesian} and found wide application in commercial soft keyboards~\cite{goodman2002language, vertanen2015velocitap}, mitigating key size constraints~\cite{vertanen2019velociwatch}.
The likelihood that the user intends to press a key is determined based on the location of a touch point on a Gaussian distribution fitted to previous observations for the respective key~\cite{goodman2002language, bi2013ffitts, rot_dual_gauss}.
For text input, the likelihood can be combined with a language prior to estimate the probability for a character in a respective location~\cite{goodman2002language}.
Based on the previous input history, the prior can be obtained through a language model, often implemented as an n-gram with finite context~\cite{vertanen2015velocitap, wordmultiplewordandsentenceinput, phraseflow} or more recently as a deep neural network~\cite{ghosh2017neural, xu2018deeptype}.
Zhu et al.~\cite{zhu2020using} propose a technique to extend this concept to general point-and-click command input by obtaining a prior based on previous selection history.



\citeauthor{rogers2010fingercloud} previously estimated the uncertainty of finger movements due to the low sensor resolution of a capacitive array via a particle filter, which they leveraged for interactive purposes~\cite{rogers2010fingercloud}.
Prior research further indicates that the display of uncertainties leads to appropriately adapted user behavior~\cite{kording2004bayesian, williamson2006continuous}.

In this paper, we propose a framework that extends probabilistic command prediction using Bayes' theorem with the aleatoric uncertainties about inferred touch locations due to sensing inaccuracies.
Thus, \projectname{} accounts for both user and sensing errors, and leverages prior context to more accurately infer intended input.
The uncertainty about the touch location guides the probabilistic framework to focus not only on a single point but consider a broader area depending on the level of uncertainty.

\subsection{Text entry systems for mixed reality}
Prior work has explored alternative text entry interfaces for MR and VR headsets to enable efficient typing without a physical keyboard.

Mid-air typing via tracked index fingers using the built-in cameras of MR/VR headsets has been demonstrated as a viable alternative to controller-based text entry~\cite{wan2024design, dudley2023evaluating} with mean text entry rates between 17.75\,WPM~\cite{dudley2018fast} and 26.1\,WPM~\cite{yi20222d} and is available on commercial devices including Quest~\cite{quest3}.
\citeauthor{markussen2014vulture}~\cite{markussen2014vulture} and \citeauthor{shen2023fast}~\cite{shen2023fast} have investigated word-gesture keyboards in mid-air for faster input.
ATK enables 10-finger typing mid-air but relies on an external LeapMotion sensor~\cite{ATK}.
Yet, mid-air interaction has been shown to lead to fatigue when used over extended periods of time~\cite{jang2017modeling}.
Eye-based text entry is another alternative but so far is limited by the text entry speeds users can achieve~(11\,WPM~\cite{cui2023glancewriter}).

To offer passive haptic feedback, researchers have leveraged body surfaces.
PinchType~\cite{pinchtype} allows users to select groups of characters by pinching with the thumb against a respective finger, which is tracked using a marker-based optical system.
Related efforts have detected finger touches using additional wearable instrumentation of the hand in various configurations, such as by appropriating one (12\,WPM)~\cite{xu2019tiptext, drgkeyboard}, both fingertips (23\,WPM)~\cite{BiTipText}, or all finger segments~\cite{liu2023printype, fingert9} as touchpads.
STAR~\cite{kim2023star} proposes a two-thumb text entry interface that leverages the hand as a surface for haptic feedback, tracks the thumbs using the cameras of the Hololens, and detects touches using capacitive tape on the fingertips, achieving a mean text entry rate of 22\,WPM. 

Similarly, flat rigid surfaces like tables can be appropriated as touch interfaces. 
QwertyRing enables one-finger typing on surfaces sensed by a finger-mounted IMU~\cite{QwertyRing}.
\citeauthor{grady2024pressurevision++}~\cite{grady2024pressurevision++} investigates index finger-based text input on a table-projected keyboard, with touch events estimated by an external static RGB camera. 

TelemetRing~\cite{telemetring} and TapStrap~\cite{tapstrap2} equip all fingers with accelerometers to identify individual fingers hitting a surface, enabling chord-based character entry.
\citeauthor{zhang2022typeanywhere}~\cite{zhang2022typeanywhere} employ a fine-tuned transformer model to decode sequences of character groups, input by users through individual finger strokes using TapStrap.
Similarly, TapType~\cite{chi2022-taptype} detects individual fingers hitting the surface via a wristband with integrated IMU sensors.
To address the greater uncertainty in finger classification, the authors propose a Bayesian neural network to predict a distribution over all ten fingers, which is then used in a probabilistic text decoder.

Prior work has demonstrated that users can transfer skills acquired on a physical keyboard to typing on flat surfaces~\cite{findlater2011typing}, with skilled typists showing spatially consistent key press distributions even during eyes-free touch typing~\cite{rashid2008relative, shi2018toast, zhu2018typing, restype}. 
To decode such surface touch typing, \citeauthor{richardson2020decoding}~\cite{richardson2020decoding} developed a neural network to directly estimate typed text from keyboard-relative spatial hand-tracking features derived from an external marker-based motion capture system.
More recently, \citeauthor{richardson2024stegotype}~\cite{richardson2024stegotype} enhanced their method to infer typed text from markerless egocentric hand tracking~\cite{han2022umetrack}, reporting a typing speed of 42.4\,WPM and 7.0\%\,UER without the aid of an additional language model.

Our method, \projectname{}, infers geometrically accurate touch locations that support broader surface-touch interactions in MR/VR.
We demonstrate the effectiveness of our method for on-surface text entry.
Instead of directly predicting intended keys from tracked hand motions, we resolve uncertainties using our probabilistic framework, which relies on inferred touch locations and priors from both character- and word-level language models.
The estimated uncertainties about touch locations aid in controlling beam search decoding~\cite{banovic2019limits}, broadening the distribution of location-based key likelihoods across the keyboard in cases of higher uncertainty.

%% file: sections/3-method.tex
\begin{figure*}[t]
    \centering
    \includegraphics[width=\textwidth]{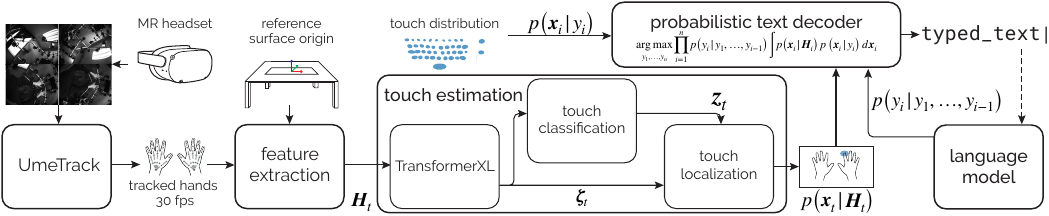}
    \caption{Overview of our framework enabling text entry on surfaces in MR. Our touch estimation network receives hand-tracking feature sequences, $\bm{H}_{t}$, as input. The features are derived from hand poses tracked from egocentric views via the headset's cameras and are normalized relative to a coordinate system anchored to a surface-aligned virtual keyboard. The network estimates the occurrence of touch events, the identity of the touching finger $\bm{z}_{t}$, and a bivariate Gaussian distribution for the touch location $p\!\left(\bm{x}_{t} \vert \bm{H}_{t} \right)$. Our probabilistic text decoder fuses this distribution with $p\!\left( \bm{x}_{i} | y_{i} \right)$—capturing the distribution of touch points for intended key presses—and a context prior based on input history from a language model $p\!\left( y_{i} | y_{1}, \ldots, y_{i-1} \right)$.}
    \label{fig:network_overview}
    \Description{This figure depicts an overview of the probabilistic text entry pipeline. The system begins with the MR headset capturing images of the user's hands from egocentric views. Using a fine-tuned pose regression network based on UmeTrack, we estimate hand poses at 30 frames per second, from which we extract a feature sequence, $\bm{H}{t}$, relative to a coordinate system anchored to a surface-aligned virtual keyboard. The first stage of the pipeline is the 'touch estimation network', where the hand-tracking features are fed into a TransformerXL network. This part of the system classifies touch events and localizes them, outputting two pieces of information: the identity of the touching finger, symbolized as $\bm{z}{t}$, and a bivariate Gaussian distribution representing the touch location, expressed as $p\left(\bm{x}{t} \vert \bm{H}{t} \right)$. The next stage is the 'probabilistic text decoder', which takes the touch location distribution and combines it with $p\left( \bm{x}{i} | y{i} \right)$, which is the distribution of touch points for intended key presses. This stage also incorporates a context prior based on the input history from a language model, denoted as $p\left( y_{i} | y_{1}, \ldots, y_{i-1} \right)$. The final output of the pipeline is 'typed_text', where the decoder has resolved the input into a coherent string of text.}
\end{figure*}

\section{Method}

In the following, we introduce our method \projectname{} for recognizing ten-finger touch input on passive surfaces based on egocentric hand tracking from an MR headset. 
Our method takes the temporal sequence of tracked hand poses as input and predicts the occurrence of a touch event, the identity of the finger making contact, and a probability distribution capturing the uncertainty about the 2D location of the touch point on the surface.
We further demonstrate how to incorporate the predicted touch location distribution into a probabilistic command prediction framework, which we implement in the form of a text entry system that enables ten-finger text input in real time on an MR headset.

\subsection{Uncertainty-aware touch estimation}

\subsubsection{Network architecture}
For hand tracking, we rely on a fine-tuned end-to-end trained 3D feature extraction and pose regression network based on \umetrack{}~\cite{han2022umetrack}.
The network takes a variable number of cropped hand images as input, depending on the number of MR headset cameras capturing the hand, and runs at 30\,Hz on Quest~3.
For each hand, the hand-tracking model's output frames consist of a joint angle vector \(\bm{\theta}\in\mathbb{R}^{20}\) representing the local pose of the hand skeleton, a 6 degrees of freedom (DOFs) global root transformation of the hand \(\bm{T}_{H}\), the user's hand scale \(S\), and a hand confidence score $\phi$.
From the hand mesh generated for each frame based on \(\bm{\theta}\), \(\bm{T}_{H}\), and \(S\) through linear blend skinning (LBS), we derive 3D key points located at the pad, edge, and tip of each finger.
These key points capture the finger's direction and volume, which is a compact yet comprehensive representation for identifying touch events, and are normalized relative to a surface-anchored world-coordinate system.
For example, in the context of text input, the normalization reference point might be the center of the keyboard between the \emph{F} and \emph{G} keys.
We stack the three 3-dimensional key points for each finger along with the confidence score $\phi$ of each hand to form a vector $\bm{h}\in \mathbb{R}^{92}$.
Our touch estimation network then receives the sequence, $\bm{H}_{t}=\{\bm{h}_{t-\tau}\}_{\tau=0}^{T-1}$, based on the last \(T\) output frames from the hand-tracking model as input.

The core of our touch estimation network is built upon an 8-layer TransformerXL model~\cite{dai2019transformer}, which uses a left-context for self-attention of 4 frames and integrates a convolutional layer with a kernel size of 3 before each attention layer, resulting in an overall context window length of 41 frames (\(T=41\)).
The output embedding $\bm{\zeta}_{t}$ for each frame at $t$ is fed to the \emph{touch classification} module consisting of one linear layer that predicts an 11-dimensional softmax output probability distribution, $\bm{z}_{t}\in\mathbb{R}^{11}$, indicating whether any of the 10 fingers of the tracked hands is making contact with the surface or whether it is a \emph{blank} frame, $\epsilon$.

The \emph{touch localization} module, consisting of two linear layers, estimates a 2D Gaussian distribution for the touch location relative to the surface-aligned x-y coordinate plane of the reference coordinate system,

\begin{equation}
p\!\left(\bm{x}_{t} \vert \bm{H}_{t} \right)=p\!\left(\bm{x}_{t} \vert \bm{\zeta}_{t} \right) = \mathcal{N}\!\left(\bm{x}_{t} \vert \bm{\mu}_{\bm{H}_t}, \bm{\Sigma}_{\bm{H}_t} \right),
\label{eq:touchposegauss}
\end{equation}
parameterized by a mean \(\bm{\mu}_{\bm{H}_t}\) and covariance matrix \(\bm{\Sigma}_{\bm{H}_t}\) for every input frame $t$.
This captures the likelihood that the user's touch occurred on a specific location $\bm{x}_t$, given the hand-tracking features \(\{\bm{h}_{t-\tau}\}_{\tau=0}^{T-1}\).
We use \(\bm{z}_{t}\) to mask the output of the predicted touch location, which we only consider in case the most likely class output of $\bm{z}_{t}$ is not the blank frame, i.e.,~$\argmax\bm{z}_{t}\neq\epsilon$.
Additionally, if the same finger class is predicted for several consecutive frames, we only consider the first frame of the sequence.
By treating the touch location as a probabilistic distribution, our approach enables the touch estimation network to account for the inherent aleatoric uncertainties due to stochasticity and limitations inherent in the input, considering multiple factors such as occlusion, temporal jitter, and biases in the tracked hand poses.

\subsubsection{Touch estimation loss.} The loss function for our touch estimation network consists of two terms for the touch classification $\mathcal{L}_{\bm{z}}$ and $\mathcal{L}_{\bm{e}}$ as well as one term for the localization $\mathcal{L}_{\bm{x}}$,

\begin{equation}
    \mathcal{L} = \alpha_{c} \mathcal{L}_{c} + \alpha_{e} \mathcal{L}_{e}  + \alpha_{x} \mathcal{L}_{x}.
\end{equation}

\paragraph{Classification loss.} For recognizing touch events, we employ the Connectionist Temporal Classification (CTC) loss~\cite{graves2006connectionist} to address imbalances related to blank frames and accommodate the sequential nature of touch events,

\begin{equation}
    \mathcal{L}_{c} = - \log \left(\sum_{\pi \in \mathcal{A}} \prod_{t} \hat{z}_{t, \pi_{t}} \right),
\end{equation}

where $\mathcal{A}$ is the set of all possible alignments between the target finger sequence and the output sequence of the touch classification module, $\pi$ represents one possible alignment, and $\pi_{t}$ is the label at position $t$ in alignment $\pi$ with $\hat{z}_{t, \pi_{t}}$ being the predicted probability for it.
This provides the network with increased temporal flexibility in deciding the optimal timing for predicting a touch event.
In addition, to ensure that the network maintains latency within acceptable bounds, we add a cross-entropy loss term, which encourages the finger classification module to predict the correct finger touch event $d$ frames following the ground-truth touch event,

\begin{equation}
    \label{eq:entropy}
    \mathcal{L}_{e} =   -  \sum_{t \in \mathcal{Z}} \sum_{j} z_{t,j} \log \left(\hat{z}_{t,j} \right).
\end{equation}

Here, $z_{t,j}$ and $\hat{z}_{t,j}$ are the $j$-th elements of the ground-truth one-hot vector $\bm{z}_{t}$ and the predicted softmax output vector $\hat{\bm{z}}_{t}$, respectively. The set $\mathcal{Z}$ includes all frames where a ground-truth contact event is detected, defined as $\mathcal{Z} = \{t \mid \argmax(\bm{z}_{(t-d)}) \neq \epsilon \}$.

\paragraph{Localization loss.} For the localization module, we minimize the negative log-likelihood (NLL) of the predicted touch distribution $p\!\left(\bm{x}_{t} \vert \bm{H}_{t} \right)$ with regard to the actual touch location $\bm{x}_{t}$.
To reduce the influence of predictive variance on the gradient calculations and thereby prevent suboptimal model fitting, we apply the $\beta$-NLL~\cite{Seitzer2022PitfallsOfUncertainty}, which weights the contribution for each sample in the loss function by its $\beta$-exponentiated variance estimates.
This loss has been developed for the univariate Gaussian NLL.
For adaptation to our scenario involving bivariate Gaussian NLL, we constrain our covariance matrix, $\bm{\Sigma}_t$, to be diagonal.
This assumption of an uncorrelated bivariate Gaussian distribution allows the loss function to be expressed as a sum of univariate Gaussian NLL terms,

\begin{equation}
    \mathcal{L}_{x} = \sum_{t \in \mathcal{Z}} \sum_{j=0}^{1} \lfloor \sigma^{2\beta}_{\bm{H}_t, j} \rfloor \left(\frac{1}{2} \log \left(\sigma^{2}_{\bm{H}_t,j} \right) + \frac{\left( x_{t,j} - \mu_{\bm{H}_t,j} \right)^2}{2 \sigma_{\bm{H}_t,j}^{2}} \right),
\label{eq:lx}
\end{equation}

where $\sigma_{\bm{H}_t, 0}$ and $\sigma_{\bm{H}_t,1}$ represent the diagonal elements of $\bm{\Sigma}_{\bm{H}_t}$; $\mu_{\bm{H}_t,0}$ and $\mu_{\bm{H}_t,1}$, and $x_{t,0}$ and $x_{t,1}$ correspond to the elements of $\bm{\mu}_{\bm{H}_t}$ and $\bm{x}_t$. The $\lfloor \cdot \rfloor$ symbol denotes the stop-gradient operator.

\subsection{Probabilistic command prediction}
In the following, we demonstrate how to integrate the uncertain prediction of a touch location into probabilistic frameworks that adopt Bayes' theorem to infer intended touch selection targets~\cite{bi2013bayesian} and commands~\cite{zhu2020using}.
While this approach, in principle, extends to other touch sensing modalities and command input tasks, we will demonstrate it for touch estimated from tracked hand poses for text input (see~\autoref{fig:network_overview}).

We consider a sequence of intended commands (in our case, text entry keys or characters), denoted as $y_{1}, y_{2}, \ldots, y_{n}$, of length $n$.
These commands correspond to touch points $\bm{x}_{1}, \bm{x}_{2},\ldots,\bm{x}_{n}$ and are associated with windows of sensor observations $\bm{H}_{1}, \bm{H}_{2}, \ldots, \bm{H}_{n}$ (in our case, tracked hand poses).
Our goal is to identify the most likely sequence of intended characters $\hat{y}_{1}, \hat{y}_{2}, \ldots, \hat{y}_{n}$ based on the observed sensor data.
We formulate this problem as

\begin{equation}
\begin{split}  
  \hat{y}_{1}, \ldots, \hat{y}_{n} &= \underset{y_{1}, \ldots, y_{n}}\argmax~p\!\left(y_{1}, \ldots, y_{n}| \bm{H}_{1}, \ldots, \bm{H}_{n}\right) \\
  &=\underset{y_{1}, \ldots, y_{n}}\argmax~p\!\left(y_{1}, \ldots, y_{n}, \bm{H}_{1}, \ldots, \bm{H}_{n}\right) \\
  &=\underset{y_{1}, \ldots, y_{n}} \argmax \prod_{i=1}^{n} p\!\left( y_{i},\bm{H}_{i} | y_{1}, \ldots, y_{i-1} , \bm{H}_{1}  \ldots, \bm{H}_{i-1}\right).
\end{split}
\label{eq:HMMproblem}
\end{equation}
Assuming that tracked hand poses across typed characters are conditionally independent given their corresponding intended key and that the prior on a character is conditionally independent given the previously typed text, we can rewrite Equation~\eqref{eq:HMMproblem} as

\begin{equation}
\begin{split}  
  \hat{y}_{1}, \ldots, \hat{y}_{n} &= \underset{y_{1}, \ldots, y_{n}}\argmax \prod_{i=1}^{n} p\!\left( \bm{H}_{i} | y_{i} \right) p\!\left( y_{i} | y_{1} \ldots, y_{i-1}  \right). \\
\end{split}
\label{eq:MLE_fact}
\end{equation}

Due to the generally greater complexity of the space of possible tracked hand poses, estimating the likelihood of a specific hand pose sequence given an intended command is challenging.
However, we can leverage the law of total probability to rewrite $p\!\left( \bm{H}_{i} | y_{i} \right)$ as

\begin{equation}
\begin{split}
p\!\left( \bm{H}_{i} | y_{i} \right) & = \int p\!\left( \bm{H}_{i} , \bm{x}_{i} | y_{i} \right) d\bm{x}_{i} \\
 & = \int p\!\left( \bm{H}_{i}  | \bm{x}_{i}, y_{i}  \right) p\!\left(\bm{x}_{i} | y_{i} \right) d\bm{x}_{i}.
\end{split}
\end{equation}

We also make the approximation that touch points $\bm{x}_{i}$ are uniformly distributed over the touch surface (i.e., $p\!\left(\bm{x}_{i}\right) \approx c, c \in \mathbb{R}$).
Additionally, we consider tracked hand pose sequences $\bm{H}_{i}$ to be conditionally independent of the intended character $y_{i}$ once the touch location $\bm{x}_{i}$ is known, simplifying to $p\!\left( \bm{H}_{i} | \bm{x}_{i}, y_{i} \right) := p\!\left( \bm{H}_{i} | \bm{x}_{i} \right)$.
This simplification ignores the character-specific nuances across hand pose sequences (i.e., distinct hand motions for different characters).
These character-specific gestures can vary significantly across individuals and are more complex to model.
By adopting this simplification, our model aims to better generalize across users.

We rewrite Equation~\eqref{eq:MLE_fact} as
\begin{equation}
\begin{split} 
    & \underset{y_{1}, \ldots, y_{n}}\argmax \prod_{i=1}^{n}  p\!\left( y_{i} | y_{1} \ldots, y_{i-1}  \right) \int p\!\left( \bm{H}_{i}  | \bm{x}_{i} \right) p\!\left(\bm{x}_{i} | y_{i} \right) d\bm{x}_{i} \\
    = &   \underset{y_{1}, \ldots, y_{n}}\argmax \prod_{i=1}^{n}  p\!\left( y_{i} | y_{1} \ldots, y_{i-1}  \right) \int 
 \frac{p\!\left(  \bm{x}_{i}  | \bm{H}_{i} \right)p\!\left( \bm{H}_{i} \right)}{p\!\left(  \bm{x}_{i}  \right)} p\!\left(\bm{x}_{i} | y_{i} \right) d\bm{x}_{i} \\
 = & \underset{y_{1}, \ldots, y_{n}}{\argmax} \prod_{i=1}^{n} \underbrace{ p\!\left( y_{i} | y_{1}, \ldots, y_{i-1} \right)}_{\text{context prior}} \int \underbrace{p\!\left( \bm{x}_{i} | \bm{H}_{i} \right)}_{\text{sensing}}\underbrace{p\!\left(\bm{x}_{i} | y_{i} \right)}_{\text{user}} d\bm{x}_{i},
 \label{eq:MLE}
\end{split}
\end{equation}

since following our assumptions $p\!\left(\bm{H}_{i}\right)$ and $p\!\left(\bm{x}_{i} \right)$ do not affect the solution. 
Equation~\eqref{eq:MLE} thus allows us to find the most likely character sequence based on a \emph{context prior} $p\!\left( y_{i} | y_{1}, \ldots, y_{i-1} \right)$ based on previous input history and the likelihood of observing a touch location given a considered key.
For this likelihood, our model accounts for \emph{both} the variation in touch location for a given key due to \emph{user} behavior and the uncertainty in the observed touch location due to the \emph{sensing} pipeline's limitations.
The context prior can be estimated by a language model.

Following prior research on ten-finger typing on touch surfaces \cite{findlater2011typing, shi2018toast} and key selection on interactive surfaces \cite{goodman2002language, bi2013ffitts}, we posit that for a given key $y$, the touch points follow a bivariate Gaussian distribution with a mean $\bm{\mu}_{y}$ and a full covariance matrix $\bm{\Sigma}_{y}$,
\begin{equation}
    p\!\left( \bm{x}_{i} | y_{i} \right) = \mathcal{N}\!\left( \bm{x}_{i} | \bm{\mu}_{y_{i}}, \bm{\Sigma}_{y_{i}} \right).
    \label{eq:touchdistribution}
\end{equation}
This distribution characterizes the touch points resulting from a user's attempt to select a key via a touch on a surface-anchored keyboard, capturing the user's mental model of the keyboard and the spatial error distribution relative to the keys' locations.
The likelihood $p\!\left( \bm{x}_{i} | y_{i} \right)$ thus relates to the uncertainty about the user's intended key given a touch location.
The uncertainty of the touch location due to the sensing or inference pipeline $p\!\left( \bm{x}_{i} | \bm{H}_{i} \right)$ weights the likelihood for a given touch location according to the likelihood that a touch occurred at this specific location.
In our case, this term is estimated by our touch estimation network (see~Equation~\eqref{eq:touchposegauss}).

Given that we integrate over the product of two bivariate Gaussian distributions (see Equations~\eqref{eq:touchposegauss}~and~\eqref{eq:touchdistribution}), there exists a closed form solution~\cite{petersen2008matrix},
\begin{equation}
\begin{split}
    \int p\!\left(\bm{x}|\bm{H}\right) p\!\left(\bm{x} | y\right) d \bm{x} 
    & = \int  \mathcal{N}\!\left(\bm{x} | \bm{\mu}_{\bm{H}}, \bm{\Sigma}_{\bm{H}}\right) \mathcal{N}\!\left(\bm{x} | \bm{\mu}_{y}, \bm{\Sigma}_{y}\right) d \bm{x} \\
    & = \rho_c \int \mathcal{N}\!\left(\bm{x}|\bm{\mu}_c, \bm{\Sigma}_c\right) d \bm{x} \\
    & =\rho_c .
\end{split}
\end{equation}

Here,

\begin{equation}
\label{eq:closedform}
\begin{gathered}
\rho_c=\mathcal{N}\!\left({\bm{\mu}_{\bm{H}}} | \bm{\mu}_{y},\left(\bm{\Sigma}_{\bm{H}}+\bm{\Sigma}_{y}\right)\right) \\
={\footnotesize \frac{1}{\sqrt{\operatorname{det}\left(2 \pi\left(\bm{\Sigma}_{\bm{H}}+\bm{\Sigma}_{y}\right)\right)}} \exp \left[-\frac{1}{2}\left(\bm{\mu}_{\bm{H}}-\bm{\mu}_{y}\right)^T\left(\bm{\Sigma}_{\bm{H}}+\bm{\Sigma}_{y}\right)^{-1}\left(\bm{\mu}_{\bm{H}}-\bm{\mu}_{y}\right)\right]} ,
\end{gathered}
\end{equation}
\begin{align}
\bm{\mu}_c &= \left(\bm{\Sigma}_{\bm{H}}^{-1}+\bm{\Sigma}_{y}^{-1}\right)^{-1}\left(\bm{\Sigma}_{\bm{H}}^{-1} \bm{\mu}_{\bm{H}}+\bm{\Sigma}_{y}^{-1} \bm{\mu}_{y}\right) , \\
\bm{\Sigma}_c &= \left(\bm{\Sigma}_{\bm{H}}^{-1}+\bm{\Sigma}_{y}^{-1}\right)^{-1} .
\end{align}

Thus, using Equation~\eqref{eq:closedform}, we can efficiently compute the combined likelihood factor.
Moreover, since these operations are fully differentiable, the resulting likelihood factor $\rho_{c}$ can be incorporated into another loss term.
This term would directly penalize the classification of a command, although this application is beyond the scope of our current work and presents an opportunity for future research to include additional motion information.

%% file: sections/4-implementation.tex
\section{Implementation}

We now describe the implementation of our interactive ten-finger text entry system that operates in real-time on a Quest~3 headset.
To implement our system, we collected an egocentric touch input dataset.
We then designed, trained, and compiled our touch estimation network, which we deployed with a text decoder following our probabilistic command prediction framework.
The decoder estimates a distribution over the keyboard keys and combines it with a language prior from an n-gram model.

\subsection{Egocentric touch input dataset}
\label{sec:touchdata}

To collect data, we recruited participants who had a baseline level of experience with touch typing on QWERTY keyboards. Participants were given a 60-second physical keyboard typing test in which they had to sustain a 45\,WPM entry rate. We collected a dataset from 385 participants, 90.3\% of whom were right-handed, 47.4\% of whom were male, and with ages ranging from 23 to 38 years old.

\subsubsection*{Apparatus} During data collection, participants' hands were adorned with 3\,mm hemisphere retro-reflective markers, sufficient to enable high-fidelity ground-truth marker-based hand tracking with an \optitrack\ motion capture system~\cite{han2018online}.
Participants wore a headset equipped with four monochrome wide-field-of-view cameras that captured synchronized VGA images at 30\,Hz.
Participants would tap their fingers on a synchronized Sensel Morph touchpad~\cite{senselmorph}, which was used to record ground-truth contact points.
OptiTrack marker trees were attached to both the Sensel Morph touchpad and the headset so that the touchpad, ground-truth hand tracking, and egocentric hand tracking could all be represented in a common coordinate space.
The 3\,mm markers used for ground-truth hand tracking are small enough not to be visibly obvious in VGA-resolution monochrome cameras mounted on a headset and thus do not interfere with egocentric markerless hand tracking. 
To motivate data collection, we tasked participants with a text transcription task. We placed a paper print-out of a QWERTY keyboard layout on top of the Sensel Morph touchpad and presented participants with prompts to transcribe on this touchpad keyboard while we recorded ground-truth 2D contact information.

\subsubsection*{Procedure} During the recording sessions, participants performed 9~blocks of transcription, with each block consisting of 5 minutes of typing phrases drawn from a single corpus.
The corpora comprised short phrases from \citeauthor{mackenzie2003phrase}'s phrase set~\cite{mackenzie2003phrase}, various short text sentences sourced from the internet, and randomly generated sequences of characters.
Participants were provided with coarse real-time feedback in the form of a cursor illustrating how much of the prompt had been typed.

\subsubsection*{Data analysis} 
In post-hoc, we corresponded the motion-capture-based hand-tracking fingertips with the Sensel Morph touchpad contact events to establish the active finger responsible for each contact. Often, there are multiple proximal contacts, which leads to ambiguity in assigning finger correspondences. To help resolve ambiguity, we applied a bipartite graph matching algorithm to establish assignments that would remain consistent over time.

Additionally, we compute the bivariate Gaussian distribution \(p(\bm{x} | y) = \mathcal{N}(\bm{x} | \bm{\mu}_{y}, \bm{\Sigma}_{y})\) from the distribution of touch points on the keyboard for each target key \(y\), and use part of the dataset to fine-tune our hand-tracking network.

\subsection{Touch estimation network}

We implemented the network in PyTorch.
Using the fine-tuned UmeTrack network, which runs in real-time on the Quest~3, we obtain the hand pose sequence from the egocentric camera footage captured by the headset.
The touch estimation network then receives the sequence $\bm{H}_{t}$, which comprises the 3D positions of key points at the pad, tip, and edge of each finger, as input.
The coordinates of these key points are relative to an origin set at the center of the keyboard between the \emph{F} and \emph{G} keys.
To minimize unseen and unrelated inputs, we further zero any key points in the input that fall outside of a bounding box surrounding the keyboard region.
After training, we compile the model using the JIT QNNPACK to run on the headset.
The network also receives and stores the touch distributions for the keyboard keys.
Equation~\ref{eq:closedform} allows for efficient computation of the combined likelihood factor $\rho_{c}$ using tensors.

\subsection{Probabilistic text decoder}

The text decoder combines the output of the touch estimation network with a language prior estimated by a language model based on the sequence of previously entered characters.
We implemented the decoder in C++.

\subsubsection*{Language model}
We deployed an n-gram language model using \emph{KenLM}~\cite{heafield-2011-kenlm}.
To obtain a prior for each entered character, we employed a 6-gram character language model with a vocabulary including all lowercase characters of the Latin alphabet, the digits 0 to 9, and punctuation marks such as comma `\emph{,}` and period `\emph{.}`.
In addition, we use a trigram word language model with a vocabulary of 100k English words~\cite{k100vocab}.
We trained the language models on Wikipedia after converting everything to lowercase and removing sentences with characters outside the specified character vocabulary.

\subsubsection*{Beam search}
Our text decoder combines the probability distribution over the keyboard keys, as determined by the touch estimation network, with the language priors as the sum of log probabilities.
It uses beam search to find the most likely combination of characters that form a word, operating with a beam width of 20 sequences and considering only word prefixes from the word vocabulary, as structured by a trie.
For each complete word, the decoder applies a word prior log probability from the word-level language model.
The beam search also accounts for omission and insertion errors through respective penalties~\cite{vertanen2015velocitap}.

\subsubsection*{Interaction vocabulary}

Users initiate the positioning of the keyboard by activating a virtual calibration button.
This action affixes the keyboard to the user's left index finger for five seconds.
Subsequently, they place their left hand flat on the surface until the countdown elapses.
Through this calibration process, users can position the keyboard to support comfortable interaction.

As users type, the per-character decoded output is displayed in a gray font to ensure visual consistency during fast input.
After a short delay, the most likely word from the beam search is presented in a white font below.
By pressing space, the suggested word is entered. To select the per-character output---allowing for the input of words outside the word vocabulary---users can pinch their thumb and middle finger together.
After deleting a character using backspace, the autocorrection is deactivated for the current word.

%% file: sections/5a-location.tex
\begin{table*}[h!]
    \centering
    \caption{Comparison of model training objectives across several metrics of interest, including mean and standard deviation of touch position error [mm], negative log-likelihood, precision, recall, and F$_{1}$-score for identifying the occurrence of a touch event and the specific finger involved, temporal offset [ms], CoER, as well as ChER using greedy or beam search decoding, factoring in (with uncertainty, w.u.) or ignoring (without uncertainty, wo. u.) estimated touch location uncertainty.}
    \vspace{-2mm}
    \resizebox{\textwidth}{!}{
    \begin{tabular}{lc|ccccccccccccccccccc} \toprule
    \multicolumn{1}{c}{\textbf{objective}} & \multicolumn{1}{c|}{$\bm{\beta}$} & \multicolumn{1}{c}{\textbf{mean pos.}$\downarrow$} & \multicolumn{1}{c}{\textbf{std. pos.}$\downarrow$} & \multicolumn{1}{c}{\textbf{NLL}$\downarrow$} & \multicolumn{3}{c}{\textbf{touch classification}} & \multicolumn{3}{c}{\textbf{finger classification}} & \multicolumn{1}{c}{\textbf{temp. off.}$\downarrow$}  & \multicolumn{1}{c}{\textbf{CoER}$\downarrow$} & \multicolumn{2}{c}{\textbf{Greedy ChER}$\downarrow$} & \multicolumn{2}{c}{\textbf{Beam ChER}$\downarrow$} \\ 

    \multicolumn{1}{c}{} & \multicolumn{1}{c|}{} & \multicolumn{1}{l}{} & \multicolumn{1}{l}{} & \multicolumn{1}{l}{} & \multicolumn{1}{c}{Prec.$\uparrow$} & \multicolumn{1}{c}{Rec.$\uparrow$} & \multicolumn{1}{c}{F1$\uparrow$} & \multicolumn{1}{c}{Prec.$\uparrow$} & \multicolumn{1}{c}{Rec.$\uparrow$} & \multicolumn{1}{c}{F1$\uparrow$} & \multicolumn{1}{l}{}  & \multicolumn{1}{c}{} & \multicolumn{1}{c}{w. u.} & \multicolumn{1}{c}{wo. u.} & \multicolumn{1}{c}{w. u.} & \multicolumn{1}{c}{wo. u.} \\ \hline

    \textbf{MSE}              & -          & 8.22            & 9.94   & -      & \textbf{.992} & .990 & \textbf{.991} & .984 & .953 & .963   & -1.11          & 31.07\%     & -       & 20.72\% & -       & 14.84\%   \\
    \textbf{pad proj.}        & -          & 15.24           & 8.42   & -      & \textbf{.992} & .990 & \textbf{.991} & .984 & .953 & \textbf{.964}   & 5.47           & 64.01\%     & -       & 61.02\% & -       & 69.78\%   \\
    \textbf{tip proj.}        & -          & 10.60           & \textbf{7.79}   & -       & .991 & \textbf{.991} & \textbf{.991} & .984 & .953 & .963  & 0.61           & 46.51\%     & -       & 42.82\% & -       & 51.29\%   \\
    \textbf{$\bm{\beta}$-NLL} & 0.0        & 11.29           &  10.99 & 3.96 & \textbf{.992} & .989 & \textbf{.991} & .984 & \textbf{.954} & \textbf{.964}  & 0.56           & 41.67\%        & 29.15\% & 30.04\% & 16.99\% & 24.71\%      \\
    \textbf{$\bm{\beta}$-NLL} & 0.5        & 8.08            & 11.66  & 3.22 & \textbf{.992} & .990 & \textbf{.991} & \textbf{.985} & .953 & .963   & 6.05           & 29.32\%       & 18.77\% & 20.35\% & 10.83\% & 14.82\%     \\
    \textbf{$\bm{\beta}$-NLL} & 0.8        & 7.49            & 11.65  & \textbf{3.13} & .991 & .990 & \textbf{.991} & .984 & .952 & .963   & \textbf{-0.26}          & 27.51\%       & 16.42\% & 17.70\% &  9.67\% & 12.83\%     \\
    \textbf{$\bm{\beta}$-NLL} & 0.9        & \textbf{6.30}   & 9.24   & 3.16  & \textbf{.992} & .990 & \textbf{.991} & .983 & .952 & .962   & 2.92           & \textbf{25.12\%}      & \textbf{14.77\%} & \textbf{15.78\%} &  8.49\% & 11.07\%    \\
    \textbf{$\bm{\beta}$-NLL} & 1.0        & 6.39            & 9.93   & 3.30  & .990 & .990 & .990 & .984 & .952 & .963 &   -0.28          & 25.30\%      & 15.29\% & 16.03\% &  \textbf{8.30\%} & \textbf{10.79\%}  \\
    
    \bottomrule

    \end{tabular}
    }
    \label{tab:off_results}
    \Description{The table provides a detailed comparison of various model training objectives against key performance metrics for egocentric touch input recognition. Metrics include the mean and standard deviation of touch position error in millimeters, negative log-likelihood, precision, recall, and F1 score for identifying the occurrence of a touch event and the specific finger involved, temporal offset in milliseconds, and contact-key error rate, as well as character error rate using greedy or beam search decoding, factoring in (with uncertainty, w.u.) or ignoring (without uncertainty, wo. u.) estimated touch location uncertainty. The training objectives are assessed for models with different beta values, indicating their impact on performance. Notably, the model with a beta value of 0.9 shows the lowest mean position error and contact-key error rate, suggesting its effectiveness in predicting touch input accurately in VR environments.}
\end{table*}

\section{Evaluation~1: Touch estimation}
\label{sec:touch_eval}

\subsubsection*{Data preparation and network training} We evaluated the accuracy of inferring touch points based on egocentrically tracked hand poses using our touch input dataset. 
We trained the network on a subset of the dataset consisting of 376~participants for a fixed number of 160~epochs, using a batch size of 64 and the Adam optimizer~\cite{kingma2014adam}.
The training dataset was composed of 5.29\,M~unique contact events.
We set \(\alpha_{c}=1\), \(\alpha_{e}=0.01\), and \(\alpha_{x}=0.001\).
For the cross-entropy loss term, we set $d=2$\,frames.
We then evaluated the trained network on a held-out test set with 9~participants, who performed 17\,K~touches per person on average.
The touches were distributed as follows across fingers: 5.12\%/5.48\% with the left/right pinky, 7.23\%/11.69\% with the left/right ring finger, 11.51\%/12.46\% with the left/right middle finger, 16.26\% / 16.82\% with the left/right index finger, 4.70\%/8.74\% with the left/right thumb.

\subsubsection*{Alternative implementations}
We evaluated the performance of our touch localization using different values for $\beta$.
In addition, we experimented with alternative implementations and loss functions.
We assessed the accuracy of touch location estimates derived from the surface-projected tip or pad key points of fingers classified as touching.
Moreover, we compared the $\beta$-NLL with a standard mean squared error loss (MSE).

\subsubsection*{Metrics}
\label{sec:locationmetrics}
We assessed the performance of our touch localization in terms of the mean and standard deviation of position error.
This error is defined by the average distance offset between the predicted mean of a touch location and the actual touch location, as captured by the Sensel Morph touchpad.
To evaluate the calibration of the estimated uncertainty, we determined the negative log-likelihood (NLL) of the predicted distribution $p\!\left(\bm{x}_{t} \vert \bm{H}_{t} \right)$, which is a lower bound for the evidence~\cite{hastie2009elements}.
Additionally, we evaluated the accuracy of our touch classification.
Specifically, we assessed accuracy across eleven classes: one for each of the ten fingers and a \emph{no-touch class} to account for missed and ghost touches (i.e., no ground-truth or predicted touch). 
Because we supervised with the CTC loss function and the model was able to learn a variable emission latency, we first aligned predicted and actual touch events in a manner similar to how the Levenshtein edit distance aligns text~\cite{mackenzie2002charerror}---by treating the sequence of finger identities like a string of characters.
We considered predictions made more than 5 frames before or 15 frames after the actual touch as incorrect.
We reported the precision, recall, and F$_{1}$-score for determining whether a touch occurred (i.e., differentiating between no-touch and touch). For correctly detected touches, we also calculated the precision, recall, and F$_{1}$-score for accurately identifying the finger that makes contact.
We also analyzed the temporal offset between a predicted touch event and the actual contact on the synchronized Sensel Morph touchpad.
Since the temporal granularity of our touch estimation was bound by the headset's hand-tracking frame rate, we reported the mean frame offset relative to the target offset of 2 frames ($d=2$ in Equation~\eqref{eq:entropy}) multiplied by the hand tracking update period (i.e., 1000/30\,ms).
Furthermore, we provided an assessment of the relative error rate between the ground-truth contact key (i.e., the key whose boundaries encompass the contact point as measured by the Sensel Morph touchpad) and the closest key based on the predicted mean touch location, which we termed the contact key error rate (CoER).

Finally, we evaluated our text decoder in an offline simulation using the touch events captured during the transcription of the MacKenzie and Soukoreff's phrase set.
Participants entered these phrases at a mean rate of 84.32\,WPM on the Sensel Morph touchpad.
We calculated the character error rate (ChER\footnote{Previous work uses the abbreviation CER~\cite{chi2022-taptype, vertanen2015velocitap, vertanen2019velociwatch}, which is easily confused with corrected error rate.}) as the Levenshtein distance between the decoded and target text, divided by the length of the target string. 
We analyzed the performance of our text decoder under two decoding strategies: \emph{greedy} per-character decoding, which used priors from the character-level language model, and \emph{beam} search decoding, which incorporated priors from both the character- and word-level language models.
For each decoding strategy, we calculated ChER in two scenarios: one considering the uncertainty about the estimated touch location, if available, and another treating the touch locations as deterministic (specifically, by setting the covariance matrix $\bm{\Sigma}_{\bm{H}}$ in Equation~\eqref{eq:closedform} to zero).

\begin{table}[t]
    \centering
    \caption{Comparison of mean touch position error across different training objectives and fingers (\textit{LT}: left thumb, \textit{LI}: left index finger, \textit{LM}: left middle finger, \textit{LR}: left ring finger, \textit{LP}: left pinky, \textit{RT}: right thumb, \textit{RI}: right index finger, \textit{RM}: right middle finger, \textit{RR}: right ring finger, \textit{RP}: right pinky).}
    \vspace{-2mm}
    \resizebox{\columnwidth}{!}{
    \begin{tabular}{lc|cccccccccc} \toprule
    \textbf{objective} & \multicolumn{1}{c|}{$\bm{\beta}$} & \textbf{LT} & \textbf{LI} & \textbf{LM} & \textbf{LR} & \textbf{LP} & \textbf{RT} & \textbf{RI} & \textbf{RM} & \textbf{RR} & \textbf{RP} \\ \hline
    \textbf{MSE} & - & 9.25 & 6.15 & 7.54 & 11.26 & 14.54 & 8.21 & 6.69 & 8.49 & 9.45 & 10.91 \\
    
    \textbf{pad proj.} & - & 13.77 & 18.05 & 13.17 & 11.08 & 10.18 & 10.34 & 19.20 & 14.24 & 16.08 & 23.86 \\
    
    \textbf{tip proj.} & - & 6.22 & 11.45 & 7.36 & 9.65 & \textbf{8.95} & 7.90 & 13.39 & 9.52 & 12.62 & 20.31 \\
    
    \textbf{$\bm{\beta}$-NLL} & 0.9 & \textbf{6.07} & \textbf{4.87} & \textbf{5.98} & \textbf{9.06} & 11.35 & \textbf{6.40} & \textbf{5.41} & \textbf{6.22} & \textbf{6.70} & \textbf{7.77} \\
    
    \bottomrule
    \end{tabular}
    }
    \label{tab:finger_results}
    \Description{This table serves as a comparison of mean touch position errors for different fingers across several inference methods and training objectives. The fingers are categorized as left thumb (LT), left index (LI), left middle (LM), left ring (LR), left pinky (LP), right thumb (RT), right index (RI), right middle (RM), right ring (RR), and right pinky (RP). Each finger's mean positional error is listed in millimeters, providing insights into which models and fingers result in more accurate touch position estimates. Notably, the beta-NLL with a beta value of 0.9 generally results in lower position errors, indicating a more accurate touch estimation based on hand tracking.}
\end{table}

\subsubsection*{Results} The results are presented in Table~\ref{tab:off_results}. 
The mean position error ranged from 6.30\,mm to 15.24\,mm, with the lowest error observed for the $\beta$-NLL method with $\beta = 0.9$.
NLL values ranged from 3.13 to 3.96, with the lowest NLL observed for the $\beta$-NLL method with $\beta = 0.8$.
Across various methods and $\beta$ values, touch event, and finger identity classification accuracy remained consistently high, with F$_{1}$-scores exceeding 0.99 and 0.96, respectively.
The temporal offset, which measures the difference between the target offset of 2~frames for predicted touch events and the actual moment of contact on the touchpad, varied from -1.11\,ms (early detection) to 6.05\,ms (late detection).
These deviations were generally less than the update period for a single frame, showcasing the high temporal precision of our trained touch estimation network.

We also reported the mean position error across different fingers in Table~\ref{tab:finger_results}.
Except for the left pinky finger, the method using $\beta$-NLL with $\beta=0.9$ achieved the lowest position error for all fingers compared to alternative methods.
The position errors for $\beta$-NLL were the lowest for the index finger (right: 5.41\,mm, left: 4.87\,mm), followed by the middle (right: 6.22\,mm, left: 5.98\,mm) and thumb (right: 6.40\,mm, left: 6.07\,mm), ring (right: 6.70\,mm, left: 9.06\,mm), and pinky (right: 7.77\,mm, left: 11.35\,mm) fingers.
We observed a similar trend for the MSE objective but generally higher errors compared to $\beta$-NLL.
The method that directly inferred the touch location based on the surface-projected tip key point of the tracked hand mesh outperformed its equivalent based on the pad key point, but on average resulted in a greater position error than the model that learned to predict the touch location using the $\beta$-NLL objective.

The impact of touch location accuracy is evident for the downstream task of text entry.
The contact key error rate was lowest at 25.30\% for the $\beta$-NLL objective with $\beta=0.9$, and highest at 64.01\% for the method utilizing pad key points.
With a greedy decoding strategy, ChER ranged from 61.02\% to 14.77\%, achieving its lowest with the $\beta$-NLL objective at $\beta = 0.9$ when factoring in location uncertainty, which improved decoding results for all uncertainty-aware models (i.e., using $\beta$-NLL).
Beam search decoding further reduced the ChER to 8.49\% for the $\beta$-NLL method with $\beta = 0.9$.

\subsubsection*{Discussion}
The $\beta$-NLL objective with $\beta=0.9$ achieved superior performance in terms of position error compared to other objectives, even outperforming the MSE model, while providing uncertainty estimates with comparably low NLL.
The model also reliably detected touch events with a minimal deviation (<\,3\,ms) from the target latency of 2~frames ($\sim$67\,ms) and an F$_{1}$-score of 0.991.
This touch event accuracy is ultimately critical for the reliable recognition of input events during typing.
Additionally, the model accurately classified the touching finger with a macro-averaged F$_{1}$-score of 0.963, a recall of 0.952, and a precision of 0.984.

The results further support a learning-based approach for estimating touch location.
Na\"ive touch localization based on specific hand key points likely struggles due to variations in finger angles during key hits and limited hand-tracking accuracy.

The higher position errors for the ring and pinky fingers are likely due to increased joint occlusions from an egocentric view, as shown in Figure~\ref{fig:teaser}.
Our network trained with $\beta$-NLL accounts for this via its estimated uncertainties.
For example, for the model with $\beta=0.9$, the touch location covariance matrices, $\bm{\Sigma}_{\bm{H}}$, associated higher mean uncertainties with touches by the left pinky finger—$\overline{\sigma_{\bm{H},0}}=11.6$ and $\overline{\sigma_{\bm{H},1}}=5.6$---which correlated with a higher mean position error of 9.9\,mm in the x-axis and 3.8\,mm in the y-axis.
In contrast, for touches with the left index finger, uncertainties were $\overline{\sigma_{\bm{H},0}}=4.6$ and $\overline{\sigma_{\bm{H},1}}=4.1$, corresponding to lower mean errors of 3.1\,mm on both axes.

Moreover, although the use of a diagonal covariance matrix for $p\!\left(\bm{x}_{t} \vert \bm{H}_{t} \right)$ (see Equation~\eqref{eq:lx}) constrains the network's expressiveness regarding the uncertainty about the touch location, we found that the error distribution between the predicted mean and the ground-truth touch location correlated only weakly across axes (\(\rho=0.15\)).

Our evaluation also highlighted the challenges of decoding intended text based solely on estimated touch locations.
With position errors of around 6\,mm---only one-third of an average finger's width---we observed an error rate exceeding 25\% in accurately identifying the ground-truth contact key from the predicted touch location.
Moreover, deviations between the intended character and the contact key, due to user errors, further complicate inference.

Our probabilistic framework benefits from incorporating language priors to infer the target phrases.
Beam search decoding further reduced character error rates compared to greedy per-character decoding.
Our evaluation also showed that the uncertainty about the estimated touch location enhances decoding performance.
The objective function with $\beta$-NLL thus not only improves the accuracy of the mean touch location---likely by avoiding overconfident predictions during periods of higher uncertainty---but the uncertainties also effectively guide the text decoder.

%% file: sections/5b-textentry.tex
\section{Evaluation~2: Online text entry}

Users adjust their behavior based on the feedback they receive from an interactive system.
For instance, to minimize typing errors, users may choose to type more slowly~\cite{banovic2013effect, banovic2017quantifying, banovic2019limits}.
This complicates testing changes to our probabilistic text decoder through offline simulation alone.
Thus, to demonstrate and evaluate our probabilistic framework end-to-end, we conducted an online text entry experiment in which participants transcribed phrases solely using egocentric hand tracking from the Quest~3.
We considered three different conditions: 1) \midair{}: state-of-the-art baseline condition where participants entered text using mid-air poke typing with their index fingers, commonly used in commercial products and research~\cite{dudley2023evaluating}, 2) \greedy{}: condition leveraging our probabilistic on-surface text entry system using greedy per-character decoding (with priors from the character-level language model), 3) \beam{}: condition using our probabilistic on-surface text entry system with per-word beam search decoding (with priors from the character- and word-level language models).
For \greedy{} and \beam{}, we used the touch estimation network with $\beta=0.9$ trained on our touch input dataset (see Section~\ref{sec:touchdata}).

\subsection{Study design}

\begin{figure}[t]
    \centering
    \includegraphics[width=\columnwidth]{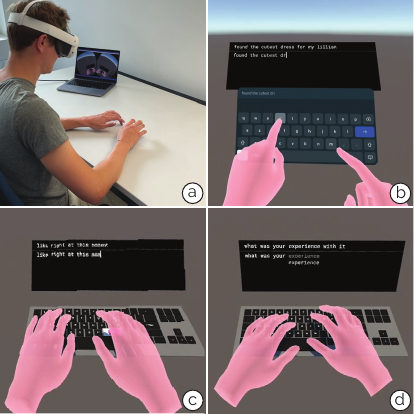}
    \caption{The figure shows the apparatus for our online text entry evaluation (a) as well as the evaluation interface for \midair{} (b), \greedy{} (c), and \beam{} (d).}
    \label{fig:studysetup}
    \Description{(a) A photo that captures a user wearing a Quest 3 headset, positioned in front of an empty desk with their hands gesturing as if typing on an invisible keyboard. (b) A screenshot showing the MIDAIR condition, where participants used mid-air poke typing with their index fingers. The VR interface consists of a virtual keyboard and a text field with an incomplete sentence, demonstrating the typing process in action. (c) A screenshot of the GREEDYSURF condition, where participants used the probabilistic on-surface text entry system with greedy per-character decoding. (d) A screenshot displaying the BEAMSURF condition, featuring our probabilistic text entry system with beam search decoding. The interface shows the target text to be transcribed and the text entered by the participant in a gray font ('exoerience') with an additional predictive text suggestion in white font ('experience') below.}
\end{figure}

\subsubsection*{Apparatus}
For the study, participants wore a Quest 3 headset.
Participants were seated in front of an empty desk. For the study setup (see~\autoref{fig:studysetup}), we implemented a virtual reality environment in Unity, featuring a virtual keyboard and a text entry field with the intended target phrase shown above.
For \midair{}, we implemented the keyboard following the \emph{Virtual Keyboard} API, which is part of the Meta XR Core SDK~\cite{virtualkeyboard}.
For \greedy{} and \beam{}, we implemented an on-surface touch keyboard following a standard QWERTY layout matching the overlay from our touch input dataset capture.
While the study was running on the headset, a moderator ensured that participants conformed to the instructions through a screenshare streamed to an external computer.

\subsubsection*{Participants}
We recruited 12~participants (5~female, 7~male, ages 23--37, mean=26.7).
On a 7-point Likert scale, all participants rated themselves 4 or higher for ``I would consider myself a fast typer'' (mean=5.5), and 4 and higher for ``I consider myself a fluent English speaker'' (mean=6.3).
The self-ratings for ``I would consider myself a touch typist'' ranged from 2 to 7 (mean=5.25, median=5.5), with two participants rating themselves below 4.
Participants self-reported their prior experience with VR technology on a 5-point Likert scale (from 1--never to 5--almost every day).
Participants' ratings ranged between 2 and 4, and their median prior experience was 3.

\subsubsection*{Task}
During each trial, participants' task was to transcribe target phrases as quickly and accurately as possible.

\subsubsection*{Procedure}
The study started with a brief introduction to the task.
Participants then completed a questionnaire, providing demographic information and self-rating their English and typing skills.
As a reference, they then transcribed 10 phrases on a physical keyboard using the TextTest++ tool, ignoring case~\cite{zhang2019beyond}.
Participants then conducted the evaluation for the three different conditions---\midair{}, \greedy{}, and \beam{}.
The order of the conditions was counterbalanced across participants.
For each condition, participants first transcribed five phrases as training, also practicing the use of selecting suggestions and deleting erroneous input.
They then completed three blocks in which they each transcribed 10 phrases from the \twitteriv{} phraseset, followed by another block with 10 phrases from the \twitteroov{} phraseset~\cite{vertanen2019velociwatch}.
The 100k English word vocabulary of our decoder included all words from the \twitteriv{} phrase set.
Each phrase from the \twitteroov{} phraseset included one out-of-vocabulary (OOV) word.
The phrases were randomly selected, ensuring that no sentence appeared twice in the study or had been used in the data collection and that they were counterbalanced across conditions.
After each condition, participants completed a NASA Task Load Index (NASA-TLX) and rated their subjective text entry speed (``I entered text quickly'' from 1--strongly disagree to 7--strongly agree) and accuracy (``I entered text accurately'' from 1--strongly disagree to 7--strongly agree) on a 7-point Likert scale.
Between blocks, participants took 1-minute breaks, and between conditions, they took 5-minute breaks.
At the end of the study, participants selected their preferred text entry method and provided qualitative feedback.

\subsection{Results}

\begin{table*}[ht]
\centering
\caption{Results of the online text entry evaluation. We reported the mean text entry rate (WPM) and error rates (UER \%, CER \%, ChER \%) across conditions (\midair{}, \greedy{}, \beam{}) and blocks (\blockone{}, \blocktwo{}, \blockthree{}, \blockoov{}). Significant pairs for post-hoc comparisons using Bonferroni correction are shown if \pvall{<}{.05}: +\,\pvall{<}{.05}, *\,\pvall{<}{.01}, **\,\pvall{<}{.001}, ***\,\pvall{<}{.0001}.}
\label{tab:onlineresults}
\resizebox{\textwidth}{!}{%
\begin{tabular}{
    l |
    S[table-format=2.2] 
    S[table-format=1.2] 
    S[table-format=2.2] 
    S[table-format=2.2] |
    S[table-format=2.2] 
    S[table-format=1.2] 
    S[table-format=2.2] 
    S[table-format=2.2] |
    S[table-format=2.2] 
    S[table-format=1.2] 
    S[table-format=2.2] 
    S[table-format=2.2] |
    S[table-format=2.2] 
    S[table-format=1.2] 
    S[table-format=2.2] 
    S[table-format=2.2]
}
\toprule
& \multicolumn{4}{c|}{\blockone{}} & \multicolumn{4}{c|}{\blocktwo{}} & \multicolumn{4}{c|}{\blockthree{}} & \multicolumn{4}{c}{\blockoov{}} \\
\cmidrule{2-17}
Condition & \multicolumn{1}{c}{WPM$\uparrow$} & \multicolumn{1}{c}{UER$\downarrow$} & \multicolumn{1}{c}{CER$\downarrow$} & \multicolumn{1}{c|}{ChER$\downarrow$} & \multicolumn{1}{c}{WPM$\uparrow$} & \multicolumn{1}{c}{UER$\downarrow$} & \multicolumn{1}{c}{CER$\downarrow$} & \multicolumn{1}{c|}{ChER$\downarrow$} & \multicolumn{1}{c}{WPM$\uparrow$} & \multicolumn{1}{c}{UER$\downarrow$} & \multicolumn{1}{c}{CER$\downarrow$} & \multicolumn{1}{c|}{ChER$\downarrow$} & \multicolumn{1}{c}{WPM$\uparrow$} & \multicolumn{1}{c}{UER$\downarrow$} & \multicolumn{1}{c}{CER$\downarrow$} & \multicolumn{1}{c}{ChER$\downarrow$}  \\
\midrule
\midair{} & 19.26 & 8.54 & \textbf{9.86} & 9.73 & 19.36 & 8.01 & \textbf{11.73} & 9.32 & 20.44 & 7.46 & \textbf{11.21} & 8.55 & 20.92 & 7.22  & \textbf{9.49}  & 8.09 \\
\greedy{}  & 35.92  & 7.21 & 16.43  & 8.53  & 34.50 & 7.56 & 19.66  & 9.19 & \textbf{37.54}  & 6.44 & 17.13  & 7.83  & \textbf{32.74} &  6.50  & 20.83  & 8.09  \\
\beam{} & \textbf{36.31}  & \textbf{2.65}& 18.14  & \textbf{3.34} & \textbf{37.32}  & \textbf{2.83}  & 18.06 & \textbf{3.62}  & 37.50 & \textbf{3.08} & 17.42  & \textbf{3.81}  & 28.27 & \textbf{3.50} & 22.70  & \textbf{4.73} \\
\hline
\hline
\multicolumn{17}{l}{WPM condition \artanova{2}{132}{72.78}{<}{.0001}: \midair{}$\overset{***}{<}$\greedy{}, \midair{}$\overset{***}{<}$\beam{}}\\
\multicolumn{17}{l}{UER condition \artanova{2}{132}{11.48}{<}{.0001}: \midair{}$\overset{*}{>}$\beam{}, \greedy{}$\overset{***}{>}$\beam{}} \\
\multicolumn{17}{l}{CER condition \anova{2}{22}{10.90}{<}{.01}{.25}: \greedy{}$\overset{*}{>}$\midair{}, \beam{}$\overset{**}{>}$\midair{}; CER block \anova{3}{33}{3.82}{<}{.05}{0.03}: \blockoov{}$\overset{+}{>}$\blockthree{}}   \\
\multicolumn{17}{l}{ChER condition \artanova{2}{132}{12.09}{<}{.0001}: \midair{}$\overset{+}{>}$\beam{}, \greedy{}$\overset{***}{>}$\beam{}}  \\
\bottomrule

\end{tabular}%
}
\Description{The table lists the results from our online text entry evaluation. It measures text entry rates in words per minute (WPM), uncorrected error rates (UER), corrected error rates (CER), and character error rates (ChER). The evaluation covers different conditions—MIDAIR, GREEDYSURF, and BEAMSURF—and blocks (BLOCK1, BLOCK2, BLOCK3, and BLOCK OOV). The aim is to understand how each text entry condition performs over time and with out-of-vocabulary phrases. The data is detailed with the mean for each metric across the conditions and blocks. Significant pairs for post-hoc comparisons using Bonferroni correction are listed at the bottom of the table. Generally, the BEAMSURF condition appears to yield the highest WPM and the lowest error rates, indicating it may offer the most efficient and accurate VR text entry method in the studied scenarios.}
\end{table*}

To assess text entry performance, we analyzed text entry rate and accuracy.
We reported the text entry rate in words per minute (WPM), determined by the time difference between the first and last keystroke for a phrase following the definition of \citeauthor{mackenzie_2015}~\cite{mackenzie_2015}.
For accuracy, we computed the uncorrected error rate (UER) and corrected error rate (CER) following TextTest++~\cite{textentrymetrics_mackenzie, zhang2019beyond}, in addition to ChER (see Section~\ref{sec:locationmetrics}). ChER is close to UER but ignores corrected characters in its computation.

\begin{figure}[ht!]
    \centering
    \includegraphics[width=\columnwidth]{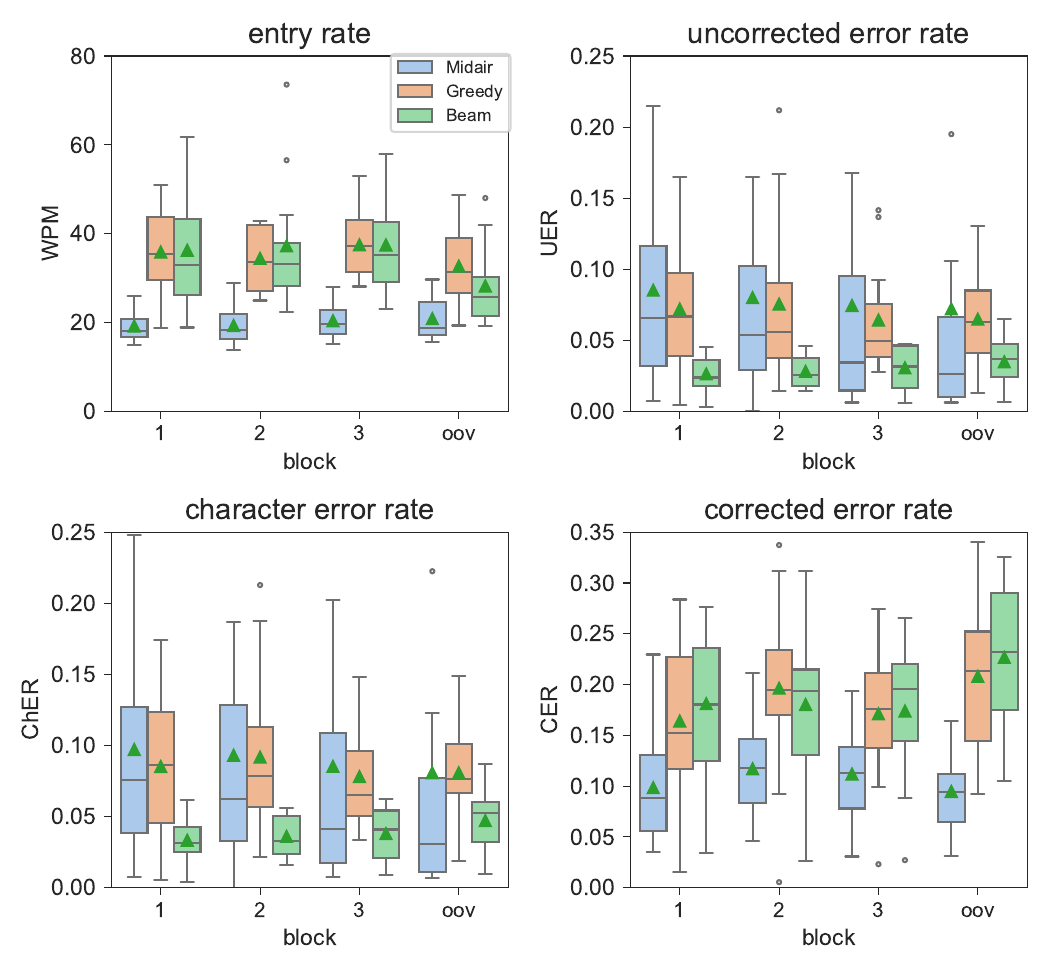}
    \caption{Boxplot of the mean text entry rate in WPM and error rates (UER, ChER, CER) across participants for \midair{}, \greedy{}, and \beam{}, and \blockone{}, \blocktwo{}, \blockthree{}, and \blockoov{}.}
    \label{fig:boxplot}
    \Description{This figure presents a detailed boxplot contrasting the mean text entry rates in Words Per Minute (WPM) and error rates (UER, ChER, CER) across different text entry methods (MIDAIR, GREEDYSURF, and BEAMSURF) over several blocks. It shows that text entry methods significantly impact both speed and accuracy, with on-surface conditions (GREEDYSURF and BEAMSURF) enabling faster text entry than MIDAIR. BEAMSURF significantly outperforms GREEDYSURF in terms of error rates. Results are also reported in Sections 6.2.1 and 6.2.2.}
\end{figure}

We reported the mean performance across participants for all conditions and blocks in~\autoref{tab:onlineresults}.
\autoref{fig:boxplot} shows the performance in terms of entry speed and error rates (UER, CER, ChER) across participants.
As a reference, on the physical keyboard, participants entered sentences with a mean speed of 67.50\,WPM (min=43.76, max=95.36, SD=15.88), a UER of~0.78\% (min=0.0, max=3.09, SD=1.01), a CER of~6.35\% (min=1.08, max=12.38, SD=3.26) and a ChER of~0.8\% (min=0.0, max=3.37, SD=1.08).

For significance testing, we considered the participant as a random factor, and the text entry condition and block as within-subject factors.
We performed a two-factor Aligned Rank Transform (ART) ANOVA for WPM, UER, and ChER as the assumptions for normality and homogeneity were not satisfied.
Post-hoc pairwise comparisons were performed using the ART-C algorithm with Bonferroni-adjusted p-values.
For CER, we performed a standard two-way repeated measures ANOVA with Greenhouse-Geisser correction as the data was normally distributed (all Shapiro-Wilk \pvall{>}{.05}) and the assumption on equal variance between groups was satisfied (all Levene's \pvall{>}{.05}).
Post-hoc pairwise comparisons were performed using Bonferroni correction.

\subsubsection{Entry rate}

\label{sec:textentryrate}
We found a main effect of the condition on WPM, but no significant difference for blocks and no significant interaction effects.
Both on-surface conditions, \greedy{} and \beam{}, allowed participants to enter text significantly more quickly than in \midair{}.
For \midair{}, the mean text entry rate was 19.26\,WPM (SD=3.60) in \blockone{}, 19.36\,WPM (SD=4.51) in \blocktwo{}, 20.44\,WPM (SD=4.17) in \blockthree{}, and 20.92\,WPM (SD=5.03) in \blockoov{} with out-of-vocabulary words.
In the \beam{} condition, participants' mean text entry rate was 36.31\,WPM (SD=14.05) in \blockone{}, 37.32\,WPM (SD=14.57) in \blocktwo{}, 37.50\,WPM (SD=11.07) in \blockthree{}, and 28.27\,WPM (SD=1.4) in \blockoov{}, while in the \greedy{} condition mean entry rate was 35.92\,WPM (SD=9.73) in \blockone{}, 34.50\,WPM (SD=7.26) in \blocktwo{}, 37.54\,WPM (SD=7.55) in \blockthree{}, and 32.74\,WPM (SD=8.79) in \blockoov{}.
We did not find any significant difference between \greedy{} and \beam{} \pval{>}{.1}.

\begin{table*}[ht]
\centering
\caption{Results from subjective ratings on speed and accuracy, and NASA-TLX task load survey for the three conditions (\midair{} (M), \greedy{} (G), \beam{} (B)) of the online text entry study. Statistical test details have been omitted for clarity. Significances for post-hoc pairwise comparisons using Bonferroni correction are indicated if \pvall{<}{.05}: +\,\pvall{<}{.05}, *\,\pvall{<}{.01}, **\,\pvall{<}{.001}.}
\label{tab:percresults}
\resizebox{\textwidth}{!}{%
\begin{tabular}{
    l |
    S[table-format=2.2]@{\,}l  |
    S[table-format=2.2]@{\,}l  |
    S[table-format=2.2]@{\,}l  |
    S[table-format=2.2]@{\,}l  |
    S[table-format=2.2]@{\,}l  |
    S[table-format=2.2]@{\,}l  |
    S[table-format=2.2]@{\,}l  |
    S[table-format=2.2]@{\,}l  |
    S[table-format=2.2]@{\,}l  
}
\toprule
  Condition & \multicolumn{2}{c|}{Speed$\uparrow$} & \multicolumn{2}{c|}{Accuracy$\uparrow$} & \multicolumn{2}{c|}{Mental$\downarrow$} & \multicolumn{2}{c|}{Physical$\downarrow$} & \multicolumn{2}{c|}{Temporal$\downarrow$} & \multicolumn{2}{c|}{Performance$\downarrow$} & \multicolumn{2}{c|}{Effort$\downarrow$} & \multicolumn{2}{c|}{Frustration$\downarrow$} & \multicolumn{2}{c}{Overall$\downarrow$} \\
\midrule
\midair{} & 2.17 & (1.34) & 3.58 & (1.51) & 59.92 & (27.20) & 67.06 & (24.15) & 52.78 & (22.92) & 58.33 & (21.40) & 68.25 & (21.23) & 61.11 & (25.83) & 61.24 & (18.33) \\
\greedy{} & 4.33 & (0.98) & 3.42 & (1.24) & 50.00 & (23.37) & 36.90 & (20.42) & 44.84 & (20.45) & 50.79 & (16.16) & 48.02 & (17.64) & 46.43 & (22.35) &  46.16 & (14.77) \\
\beam{} & \textbf{4.50} & (1.38) & \textbf{4.50} & (1.57) & \textbf{48.41} & (23.84) & \textbf{33.33} & (21.58) & \textbf{43.25} & (19.94) & \textbf{37.30} & (20.29) & \textbf{47.62} & (22.79) & \textbf{42.86} & (20.41) & \textbf{42.13} & (15.81)  \\
\midrule
Significant pairs  & \multicolumn{2}{c|}{M$\overset{**}{<}$B, M$\overset{**}{<}$G} & \multicolumn{2}{c|}{n/a} & \multicolumn{2}{c|}{M$\overset{+}{>}$B, M$\overset{*}{>}$G} & \multicolumn{2}{c|}{M$\overset{**}{>}$B, M$\overset{*}{>}$G} & \multicolumn{2}{c|}{n/a} & \multicolumn{2}{c|}{n/a} & \multicolumn{2}{c|}{M$\overset{+}{>}$B, M$\overset{*}{>}$G}& \multicolumn{2}{c|}{n/a} & \multicolumn{2}{c}{M$\overset{*}{>}$B, M$\overset{*}{>}$G} \\
\bottomrule
\end{tabular}
}
\Description{The table summarizes subjective ratings in terms of mean and standard deviation across participants on speed, accuracy, and the dimensions of the NASA-TLX survey, for the three different conditions of our online text entry study---MIDAIR, GREEDYSURF, and BEAMSURF. Significant pairs for post-hoc comparisons using Bonferroni correction are listed at the bottom of the table.}
\end{table*}

\subsubsection{Error rate}
\label{sec:texterrorrate}
There is a significant main effect of the condition on UER, but no interaction effects and main effect for blocks.
\beam{} had significantly lower UERs compared to \greedy{} and \midair{}, with no significant difference between \midair{} and \greedy{}.
We made the same observations for ChER with a significant main effect for condition, and 
a significantly lower ChER for \beam{} compared to \greedy{} and \midair{}.

For \midair{}, participants achieved a mean UER of 8.54\% (SD=6.98) and ChER of 9.73\% (SD=7.72) in \blockone{}, 8.01\% UER (SD=7.79) and 9.32\% ChER (SD=9.10) in \blocktwo{}, 7.46\% UER (SD=9.95) and 8.55\% ChER (SD=11.2) in \blockthree{}, and 7.22\% UER (SD=10.76) and 8.09\% ChER (SD=11.76) for \blockoov.
UER and ChER for \greedy{} were 7.21\% (SD=5.23) and 8.53\% (SD=5.8) in \blockone{}, 7.56\% (SD=5.95) and 9.19\% (SD=5.93) in \blocktwo{}, 6.44\% (SD=3.91) and 7.83\% (SD=4.05) in \blockthree{}, and 6.50\% (SD=3.59) and 8.09\% (SD=3.82) in \blockoov{}.
The autocorrection in \beam{} led to significantly fewer uncorrected errors with a mean UER of 2.65\% (SD=1.27) in \blockone{}, 2.83\% (SD=1.21) in \blocktwo{}, 3.08\% (SD=1.56) in \blockthree{}, and 3.50\% (SD=1.66) in \blockoov{}.
Similarly, mean ChER was 3.34\% (SD=1.56) in \blockone{}, 3.62\% (SD=1.50) in \blocktwo{}, 3.81\% (SD=1.84) in \blockthree{}, and 4.73\% (SD=2.26) in \blockoov{}.

Participants had to backspace to correct mistyped characters, requiring the deletion of correctly entered characters to correct errors at the beginning of a word and causing relatively large CER.
We found a main effect of condition and block on CER, but no interaction effects.
Pairwise comparisons revealed significant differences between \blockthree{} and \blockoov{} as well as between both \greedy{} and \beam{} compared to \midair{}.

Due to the slower entry rate, participants were likely more hesitant to perform and retype mistyped text, leading to a lower mean CER of 9.86\% (SD=5.69) in \blockone{}, 11.73\% (SD=4.85) in \blocktwo{}, 11.21\% (SD=4.90) in \blockthree{}, and 9.49\% (SD=4.37) in \blockoov{} compared to \greedy{} with 16.43\% (SD=8.36) in \blockone{}, 19.66\% (SD=9.01) in \blocktwo{}, 17.13\% (SD=6.90) in \blockthree{}, and 20.83\% (SD=8.30) in \blockoov{} and \beam{} with 18.14\% (SD=7.43) in \blockone{}, 18.06\% (SD=8.11) in \blocktwo{}, 17.42\% (SD=6.98) in \blockthree{}, and 22.7\% (SD=7.05) in \blockoov{}.

\subsection{Perceived performance and workload}

We tested for significant differences between conditions regarding perceived speed, accuracy, mental task load, physical task load, temporal task load, performance, effort, frustration, and overall task load according to the raw NASA-TLX survey (scale from 1--100).
After checking for normality and homogeneity of variances, we performed a one-way ANOVA with post-hoc pairwise comparisons using Bonferroni correction for accuracy, mental task load, physical task load, temporal task load, performance, effort, frustration, and overall task load.
For perceived speed, we conducted a one-way ART ANOVA with post-hoc pairwise comparisons using the ART-C algorithm with Bonferroni-adjusted p-values.
The mean and standard deviations across participants are reported in \autoref{tab:percresults}.

\subsubsection{Perceived speed and accuracy}
Participants perceived themselves to be significantly faster with \greedy{} and \beam{} compared to \midair{}.
Participants' mean for speed was 4.50 (SD=1.38) for \beam{}, 4.33 (SD=0.98) for \greedy{}, and 2.17 (SD=1.34) for \midair{}.
We could not find a significant difference between \beam{} and \greedy{}, which is in alignment with measured WPM (see Section~\ref{sec:textentryrate}).
Even though we observed significantly smaller UER for \beam{} (see Section~\ref{sec:texterrorrate}), participants did not perceive a significant difference in terms of accuracy between conditions.
Participants rated their accuracy on average with 4.50 (SD=1.57) for \beam{}, 3.42 (SD=1.24) for \greedy{}, and 3.58 (SD=1.51) for \midair{}.

\subsubsection{(Raw) NASA-TLX}
Mental task load, physical task load, effort and overall task load were significantly lower for \beam{} compared to \midair{}.
Similarly, mental task load, physical task load, effort, and overall task load were significantly lower for \greedy{} compared to \midair{}.
We did not find significant differences in temporal task load and frustration, nor in performance, according to the pairwise post-hoc tests between conditions after adjusting p-values using Bonferroni (all \pvall{>}{0.05}).

\subsubsection{Qualitative feedback}
Out of 12 participants, 9 participants had a preference for \beam{}, 2 for \greedy{}, and 1 for \midair{}. 
When asked about their experience, \textit{Participant~9} stated that they found \midair{} to be ``exhausting'' and they were missing ``haptic feedback''.
\textit{Participant~3} explained their preference for \midair{} due to a higher level of control, even though it involved trading off speed.
\textit{Participant~7} found typing with autocorrection in \beam{} ``much easier'', and \textit{Participant~6} specified that ten-finger typing on a surface was very intuitive, requiring them not to look at their hands while typing, similar to ``on a physical keyboard''.
\textit{Participant~1} was further impressed by the speed at which they could enter text on the MR headset.
There were also suggestions for improvements.
\textit{Participant~8} would have preferred the autocorrect for a word to remain active even after pressing backspace.

\subsection{Discussion}

Our findings indicate that both on-surface conditions outperform mid-air input in terms of text entry rate, which is also reflected in participants' subjective ratings on input speed.
The mean text entry rate of 37\,WPM for \beam{} also outperforms most MR/VR text entry methods from prior work~\cite{kim2023star}, in particular approaches based on head-mounted cameras (e.g., compare \citeauthor{yi20222d}'s 26.1 WPM~\cite{yi20222d}).
This is likely due to the passive haptic feedback, which provides users with more tangible cues compared to visual feedback alone.

The results also show the benefits of our probabilistic framework, foremost enabling text entry for both on-surface conditions based on tracked hand poses.
\beam's autocorrection leveraged ambiguous input and uncertainties over the entire word, which led to significantly smaller uncorrected error rates (<\,3\%).
These rates approach results reported for fast text input on smartphones~\cite{vertanen2015velocitap}.

Corrected error rates of \greedy{} and \beam{} indicate room for improvement in terms of accuracy, as well as the need for more efficient correction techniques that do not require the deletion of correctly decoded input. 

While we did not find a significant difference between text entry rates and uncorrected error rates across blocks, the mean entry rates for \beam{} are notably lower on phrases with OOV words (28.27\,WPM), likely because participants started to trust the autocorrection and had to perform costly corrections in case of mispredictions. 
Future iterations should focus on more versatile language models and the option for overwriting the autocorrection post-hoc.

The overall task load, according to the NASA-TLX survey, was significantly smaller for \beam{} and \greedy{} than for \midair{}.
We attribute this to significantly smaller physical demands in both conditions.
Even though participants could support their elbows on the desk during mid-air input, additionally resting their wrists on the surface allowed for even more comfortable input, which is in line with previous findings~\cite{ismar2022-comfortableUIs}.

The top-performing participant achieved a mean text entry rate of 73.6\,WPM with a UER of 1.5\% and a CER of 2.6\% in \blocktwo{} using \beam{}.
This highlights the potential of our text entry system to support fast text input for expert users.

However, we also acknowledge that there remains a gap between reference text entry rates on a physical keyboard compared to \beam{}.
Participants entered text 45\% slower on average (67.5\,WPM vs. 37.0\,WPM), indicating the need for further improvements to fully substitute physical keyboards in MR/VR environments.

%% file: sections/7-limitations.tex
\section{Limitations and Future Work}

While our evaluation provided support for our approach and confirmed the promising performance of  ten-finger touch sensing from egocentric vision using our method, several limitations motivate interesting avenues for future work.

\paragraph{Constrained Character Set}

Our method was evaluated using a keyboard with a limited set of keys, specifically excluding capitalization and function keys, such as shift.
However, our method estimates touch locations in a manner that is agnostic to specific key locations and, thus, in principle, is applicable to the entire keyboard space.
Future work should focus on expanding our approach to encompass a broader set of keys, either through the acquisition of more comprehensive touch data or by experimenting with default distributions for unseen command fields.

\paragraph{Stateful touch}
Our current touch estimation network accurately predicts the onset of touch events, including time, location, and the responsible finger, but falls short in recognizing sustained contact.
Future work could experiment with refining the network to predict touch states---press, hold, and release---using additional motion and visual finger features~\cite{grady2022pressurevision}.

\paragraph{No eye tracking} 

During the collection of the touch input dataset (Section~\ref{sec:touchdata}), participants interacted with a visible keyboard overlay.
No specific instructions were provided regarding whether participants should maintain visual contact with their hands while typing.
Consequently, the touch distribution encompasses inputs reflecting varying degrees of attentiveness and precision.
This includes more accurate input during periods of heightened attention and careful typing as well as less precise input during rapid, eyes-free typing.
Future iterations of our probabilistic text decoder could account for this by employing adapted touch distributions.
Such adaptations could be personalized and contingent upon the user's current focus of attention, approximated by the direction of gaze (many of today's headsets are already equipped with eye trackers).

\paragraph{Resolving uncertainty through motion cues}

Our current framework infers user input based on touch locations relative to command interfaces (e.g., buttons and keys).
Our approach thus generalizes across a wider range of devices and input tasks.
However, in certain tasks, including text entry, hand motions offer additional cues that could help resolve input uncertainty.
For instance, touch typists consistently use specific fingers for certain characters.
Future adaptations specifically designed for text entry could integrate hand motion information to enhance accuracy.
This kind of information may vary from one user to another, making it suitable for personalizing a model to each individual user's typing style.

%% file: sections/8-conclusion.tex
\section{Conclusion}

We have presented \projectname, a novel method for detecting touch input on physical surfaces with uncertainty based on egocentric vision from head-worn devices.
\projectname\ incorporates a neural network to accurately recognize the moment of touch events, the identity of the touching finger, and the input location on the surface.
The key novelty of our method is to explicitly integrate the uncertainties that stem from the two errors involved during detection:
user error---introduced by potential occlusion, finger softness, and motor inaccuracy---and sensing error---introduced by egocentric vantage points that lead to self-occlusion in addition to regular sensor noise.
Our network accounts for the sensing uncertainties and estimates a bivariate Gaussian distribution for the touch location.
In our evaluation, our method inferred locations with a mean position error of 6.3\,mm.

Beyond the mere uncertainty scores for estimated touch locations, we demonstrate how these uncertainties provide a key benefit as part of a probabilistic framework to decode rapid surface touch input from all ten fingers during dexterous text entry.
To enhance accuracy, our probabilistic framework also incorporates priors from both character- and word-level language models.
We evaluated our probabilistic framework as part of an end-to-end text entry system.
Participants entered text with a mean entry rate of 37.0\,WPM with 2.9\% uncorrected error rate, outperforming a mid-air keyboard baseline in performance, task load, and user preference.

Taken together, we believe that our approach enables better interaction in MR, particularly during prolonged productivity tasks that often require efficient text input in addition to direct interaction. 
Because our framework can in principle generalize to a wide range of command prediction tasks, we believe that our probabilistic and uncertainty-aware approach holds substantial promise to support all on-surface input interactions in MR in the future.